\documentclass[10pt,twocolumn,letterpaper]{article}

\usepackage{wacv}              % To produce the CAMERA-READY version
\usepackage{graphicx}
\usepackage{amsmath}
\usepackage{amssymb}
\usepackage{booktabs}
\usepackage{bbm}
\usepackage{soul}

\usepackage[pagebackref,breaklinks,colorlinks]{hyperref}

\usepackage[capitalize]{cleveref}
\crefname{section}{Sec.}{Secs.}
\Crefname{section}{Section}{Sections}
\Crefname{table}{Table}{Tables}
\crefname{table}{Tab.}{Tabs.}

\def\wacvPaperID{1167} % *** Enter the WACV Paper ID here
\def\confName{WACV}
\def\confYear{2024}

\begin{document}

\title{Deep Active Learning: A Reality Check }

\author{Edrina Gashi\\
Independent Researcher\\
\and
Jiankang Deng\\
Huawei Noah's Ark\\
\and
Ismail Elezi\\
Huawei Noah's Ark\\
}
\maketitle

%%%%%%%%% ABSTRACT
\begin{abstract}
We conduct a comprehensive evaluation of state-of-the-art deep active learning methods. Surprisingly, under general settings, no single-model method decisively outperforms entropy-based active learning, and some even fall short of random sampling. We delve into overlooked aspects like starting budget, budget step, and pretraining's impact, revealing their significance in achieving superior results.
Additionally, we extend our evaluation to other tasks, exploring the active learning effectiveness in combination with semi-supervised learning, and object detection. Our experiments provide valuable insights and concrete recommendations for future active learning studies. By uncovering the limitations of current methods and understanding the impact of different experimental settings, we aim to inspire more efficient training of deep learning models in real-world scenarios with limited annotation budgets. This work contributes to advancing active learning's efficacy in deep learning and empowers researchers to make informed decisions when applying active learning to their tasks.
\end{abstract}

\section{Introduction}
\begin{flushright}
\emph{All \st{animals} samples are equal, 
but some are more equal than others. \cite{Orwell}}
\end{flushright}
Deep Learning models have achieved a remarkable feat by attaining near-human accuracy in a multitude of tasks. This prowess is attributed to their ability to harness extensive datasets, enabling them to conquer challenges like image classification \cite{DBLP:conf/cvpr/HeZRS16}, object detection \cite{DBLP:conf/nips/RenHGS15}, and image segmentation \cite{DBLP:conf/eccv/ChenZPSA18}. These accomplishments primarily reside within the domain of supervised learning, where the models rely on vast repositories of labeled data such as ImageNet \cite{DBLP:journals/ijcv/RussakovskyDSKS15} or MS-COCO \cite{DBLP:conf/eccv/LinMBHPRDZ14}. While sourcing image data might seem straightforward, the process of meticulously annotating them is a laborious, time-consuming, and error-prone task. This challenge becomes even more pronounced in specialized fields like medical imaging, or forensics, where data annotation necessitates expert skills and can lead to significant errors.

% active learning is not the only game in town
Recent empirical investigations \cite{DBLP:conf/eccv/JoulinMJV16,DBLP:conf/eccv/MahajanGRHPLBM18} have unveiled an intriguing aspect of deep neural models' performance – it is not yet saturated concerning the size of the training data. The adage "more data, more accuracy" holds true, but it comes at a cost. This realization has propelled researchers to explore a semi-supervised approach \cite{DBLP:journals/pami/MiyatoMKI19,DBLP:conf/nips/OliverORCG18}, which seeks to strike a balance between labeled and unlabeled data. However, despite these efforts, the performance of semi-supervised learning models still lags behind their fully-supervised counterparts \cite{DBLP:conf/nips/RasmusBHVR15}.
Regardless of whether a learning scenario is fully-supervised or semi-supervised, a universal constraint persists: the annotation budget is finite. Hence, the imperative to adopt an intelligent labeling strategy – often referred to as \textit{active learning} – becomes evident. This approach involves selectively labeling the most informative samples, a technique that has been shown to significantly enhance the performance of both fully-supervised \cite{DBLP:conf/cvpr/BeluchGNK18,DBLP:conf/cvpr/YooK19,DBLP:journals/corr/abs-1904-00370} and semi-supervised \cite{DBLP:conf/eccv/GaoZYADP20} models.

The landscape of research in this domain is marked by a surge of recent papers proposing novel methods, each vying to establish its state-of-the-art credentials. However, a common theme emerges – these proposed methods often carry certain shortcomings. Some employ testing sets for validation, others make methodological testing errors, and unfair comparisons are not uncommon. Though such issues are not unique to active learning, analogous subfields have witnessed the emergence of works aimed at rectifying these pitfalls. For instance, in metric learning, \cite{DBLP:conf/eccv/MusgraveBL20} conducted an exhaustive study of various cutting-edge algorithms, revealing that the latest methods only marginally outperform classical techniques like contrastive and triplet loss in most scenarios. This scrutiny has catalyzed more meticulous research and rigorous evaluations, thereby fostering advancements in the field.

The present study embarks on a comprehensive exploration of several renowned deep active learning methods. In the experimental design, we subject these methods to uniform conditions across multiple data splits and datasets, thereby ensuring a fair comparison. Intriguingly, our findings challenge the supremacy of claimed state-of-the-art techniques. Contrary to expectations, the results reveal that, in a general setting, none of the proposed methods decisively outperform active learning based on entropy. Furthermore, some of these methods fail to consistently surpass the performance of even random sampling. This research delves further by investigating overlooked parameters such as the starting budget, budget step, and the impact of pretraining on the methods' efficacy.

To comprehensively assess the landscape, we also examine the benefits of active learning in the semi-supervised learning setting, showing that a combining these two techniques leads to better results than either of them in isolation. We then extrapolate these findings to another pertinent task – object detection. We finish this study by offering valuable insights and recommendations for future active learning endeavors, drawing parallels with the evolution of research in related subfields.

Our \textbf{contribution} is the following:

\begin{itemize}
    \item \textbf{Thorough Evaluation of Active Learning Methods}: We conduct an exhaustive and unbiased assessment of various cutting-edge active learning techniques. The results unveil a significant insight: in a general context, no single-model approach can outperform the \textit{entropy}-based strategy.
    \item \textbf{Expanded Experimental Scope}: Our investigation extends beyond the mainstream by examining the behaviors of active learning methods across varying starting budgets, budget step sizes, and their performance in conjunction with pre-trained models. This broader exploration offers a more comprehensive understanding of the methods' dynamics.
    \item \textbf{Other modalities}: We investigate the interplay between active learning and semi-supervised learning, shedding light on their synergistic potential. Furthermore, we extend our study to the realm of object detection, showcasing the applicability of the methods in different contexts.
    \item \textbf{Guidelines for Future Work}: Our research does not conclude with results; we distill our findings into actionable recommendations. These insights provide a roadmap for upcoming research endeavors in the domain of active learning, aiding researchers in charting a more effective course of exploration.
\end{itemize}

In essence, our work significantly enriches the understanding of active learning's intricacies and its potential across various dimensions, ensuring that future advancements in this field are better informed and more impactful.

\section{Related Work}

\textbf{Active learning (AL)} has garnered extensive attention over the past two decades. It centers on selecting uncertain samples for classifier prediction or those where independent classifiers disagree. A comprehensive survey \cite{DBLP:series/synthesis/2012Settles} aptly delves into this issue, particularly within the context of low-level data. The survey covers a spectrum of active learning approaches, encompassing uncertainty-based \cite{DBLP:conf/icml/LewisC94,DBLP:conf/sigir/LewisG94,DBLP:conf/ecml/RothS06,DBLP:conf/emnlp/SettlesC08,DBLP:conf/nips/LuoSU13}, SVM-based \cite{DBLP:journals/jmlr/TongK01,DBLP:conf/cvpr/VijayanarasimhanG11a}, and query-by-committee methods \cite{DBLP:conf/colt/SeungOS92,DBLP:conf/ipmi/IglesiasKMTC11}.

\textbf{Deep Active Learning (DAL)} has garnered substantial attention in recent years, resulting in a multitude of approaches addressing the problem from diverse angles. One notable strategy, presented in \cite{DBLP:conf/cvpr/BeluchGNK18}, involves training an ensemble of neural networks followed by the selection of samples exhibiting the highest acquisition scores. These scores are determined through acquisition functions such as entropy \cite{DBLP:journals/sigmobile/Shannon01} or BALD \cite{DBLP:journals/corr/abs-1112-5745}. Concurrent research \cite{DBLP:conf/icml/GalIG17,DBLP:conf/nips/KirschAG19} delves into similar territory, approximating uncertainty by leveraging Monte-Carlo dropout \cite{DBLP:conf/icml/GalG16}.
A comparative analysis performed in \cite{DBLP:conf/cvpr/BeluchGNK18} sheds light on these approaches, resolutely concluding that the ensemble-based methodology yields superior results albeit with increased computational demands. An alternate Bayesian approach is presented by \cite{DBLP:conf/icml/TranDRC19}, wherein data augmentation is combined with Bayesian networks. The authors employ a variational autoencoder (VAE) \cite{DBLP:journals/corr/KingmaW13} on real and augmented samples, selecting unlabeled samples based on the VAE's reconstruction error. An analogous strategy is outlined in \cite{DBLP:journals/corr/abs-1904-00370}, wherein a latent space is learned using a VAE in tandem with an adversarial network trained to distinguish between labeled and unlabeled data. The VAE and adversarial network engage in a minimax game, with the former attempting to deceive the latter into classifying all data points as labeled, while the adversarial network refines its ability to discern dissimilarities within the latent space.
A divergent approach is presented in \cite{DBLP:conf/iclr/SenerS18}, wherein the active learning challenge takes the form of core-set selection. This entails the identification of a subset of points such that a model trained on this subset remains competitive when applied to the broader dataset.
Another unique perspective is offered by the work of \cite{DBLP:conf/cvpr/YooK19}. Here, the authors present a heuristic yet elegant solution: a network is trained for classification while concurrently predicting cross-entropy loss. During the sample acquisition phase, samples with the highest prediction loss are deemed the most intriguing and subsequently earmarked for labeling.

\textbf{Deep Semi-Supervised Actove Learning (DSSL).} constitutes a profound approach in deep learning that merges a limited set of labeled data with a substantial pool of unlabeled data during neural network training. This contrasts with active learning (AL), where the utilization of unlabeled data is typically restricted to the acquisition phase. Within semi-supervised learning (SSL), these unlabeled data play a role throughout the training process. Several methods have demonstrated exceptional outcomes \cite{pseudo-labelling,DBLP:conf/nips/TarvainenV17,DBLP:conf/iclr/LaineA17,DBLP:journals/pami/MiyatoMKI19} by framing semi-supervised learning as a regularization challenge. This involves introducing an additional loss component for the unlabeled samples, effectively enhancing the learning process.
Subsequent endeavors have significantly advanced SSL's performance in object classification \cite{DBLP:conf/nips/OliverORCG18, DBLP:conf/nips/BerthelotCGPOR19, DBLP:conf/iclr/BerthelotCCKSZR20, DBLP:conf/nips/SohnBCZZRCKL20, DBLP:journals/corr/abs-2110-08263, DBLP:journals/corr/abs-2106-04732}. These efforts have contributed to refining the SSL paradigm and achieving remarkable results across diverse applications.

Interestingly, despite the apparent overlap between active learning and semi-supervised learning, the fusion of these methodologies has been a relatively unexplored territory. A pioneering attempt to unify these concepts surfaced in \cite{DBLP:conf/eccv/GaoZYADP20}. This work employs a consistency-based semi-supervised learning technique during training, establishing a connection between these two pivotal domains. The work was later extended in the domain of object detection \cite{al-ssl-od} giving similar conclusions.

\section{Representative Active Learning works}
\textbf{Notation:} Let $D$ be a dataset divided into a labeled set $L$ and a pool of unlabeled data $U$. 
Each sample in the dataset belongs to a class $y$, and in total there are $c$ classes.
The Active Learning acquisition function consists of mining a subset of samples from the pool of unlabeled data $U$ and transferring them to the labeled set $L$, incurring a labeling cost. 
For a sample $x$ (e.g., an image), a neural network $\theta$ generates a feature vector $f$ and a softmax probability distribution $p_i$, where $p$ represents the likelihood of the sample belonging to class $i$.
We define the labeling budget as $b$ and typically $b << |L|$. The training procedure is done for $n$ active learning cycles, and in each cycle, we label $b/n$ the most promising samples. 

\textbf{Active Learning methods:} In our framework, we consider the following representative active learning works: random, entropy, variational ratio, Bayesian Active Learning (BALD), Core-Set, and Learning Loss for Active Learning (LLAL).

\textit{Random} a(x) = unif() with
unif() is a function returning a draw from a uniform distribution over the interval [0, 1]. Using this acquisition function is equivalent to choosing a point uniformly at random from the pool. It is the default baseline in active learning.
     
\textit{Entropy} is an active learning method, where the classifier trained in this current iteration computes the softmax predictions of the unlabeled samples, and we choose to label the samples with the highest entropy (on the softmax predictions). 
The entropy of a sample is computed as:

\begin{equation}
H(x) = \sum_{i=1}^c p_i log(p_i).
\end{equation}

\textit{Variation Ratio} describes the lack of confidence of a classifier. It is computed as:

\begin{equation}
vr(x) = 1 - max(p|x).
\end{equation}

\textit{BALD} scores a data point based on how well the model’s predictions inform us about the model parameters $\theta$. For this, it computes the mutual information $\mathbbm{I}(y|\theta)$. The formula is given as:

\begin{equation}
B(x) = H(y|x) - \mathbbm{E}_{p(\theta)}[H[y|x, \theta]] = \mathbbm{I}(y, \theta|x).
\end{equation}

where $\mathbbm{E}$ represents the expectation. 

\textit{LLAL} divides a minibatch of size $B$ into $B/2$ data pairs $(x_j, x_k)$.
Then, it learns the loss prediction module by considering the difference between a pair of loss predictions, which completely makes the
loss prediction module discard the overall scale changes. For each pair, the loss function of the loss prediction module is defined as:

\begin{equation}
L_{loss}(\hat{l}, l) = max(0, \mathbbm{1} (l_k, l_j) \cdot (\hat{l_k} - \hat{l_j}) + \xi)
\end{equation}

where $\xi$ is a pre-defined positive margin, and $\mathbbm{1}$ is an indicator variable that takes values $+1$ if $l_i > l_j$, and $-1$ otherwise. Then, during inference, only the learned loss $L_{loss}$ is predicted, with the labeled samples being those with the highest $L_{loss}$.

\textit{Core-Set} divides the labeled pool into k clusters, in such a way as to maximize the spread of the labeled data. It uses a complex optimization procedure based on Gurobi \cite{gurobi} optimizer.

All the mentioned methods, except Core-set, compute the acquisition function and then choose to label the $b/n$ samples with the highest acquisition score. For the most part, these methods are also known as \textit{uncertainty} methods.
In the case of Core-set, it chooses the $b/n$ samples that maximize the representation of the data, and it is known as a \textit{diversity} method.

\subsection{Flaws in the methodology}
Active Learning research is not very standardized. 
Different methods use different backbones and typically minimal comparison with other methods.
Furthermore, most of the methods do not use a validation set, and perform experiments in only one or two datasets, usually in simple ones like CIFAR-10.
Additionally, most methods show experiments in a very limited setting (e.g., very few AL cycles).
LLAL \cite{DBLP:conf/cvpr/YooK19} in particular improves the backbone to be higher-performing for low-resolution images, and achieves better results than the other methods.
However, while we were able to reproduce their results, the results of entropy-based AL were higher than those reported in the paper, and actually similar to those of LLAL.
Other methods like Variational Adversarial Active Learning (VAAL) \cite{DBLP:journals/corr/abs-1904-00370}, as can be seen from the official code, by mistake, report the results in the training set.
Our experiments in the method show that it does not outperform random acquisition.
The Power of Ensembles method \cite{DBLP:conf/cvpr/BeluchGNK18}, while reaching very high results, comes with a significantly higher computational cost.  It also is unclear if the improvement comes from ensembles improving AL, or from the higher computational cost.

In this work, we strive for simplicity. We train all methods under the same hyperparameter configuration, seed, backbone, and training tricks.
We also use a larger number of datasets, being diverse in their complexity and image resolution.
Furthermore, we apply a larger number of AL cycles, to have more complete results.

\section{Experiments}

\subsection{Experimental setup}

We perform our experiments in four standard classification benchmarks: CIFAR-10, CIFAR-100 \cite{cifar10}, Caltech-101, and Caltech-256 \cite{caltech256}. 
We keep a unified setting, where we keep the same hyperparameters for all datasets. Changing the hyperparameters for every dataset, while it can come with a slight improvement, is unrealistic in the active learning setup.
Based on the dataset size, we do a different number of active learning cycles.
We add $1000$ images for labeling in each cycle.
We train each network from scratch at every active learning cycle.
We report the mean and standard deviation based on the training of five trials.
For each method, we use the same initial split and random seed.
We give the main hyperparameters for each dataset in Table \ref{hyperparameters}. We also provide the exact numbers (mean and standard deviation) in the supplementary.

\begin{table}[]
\begin{tabular}{l|c|c|c|c}
                      & \multicolumn{1}{l}{CIF10} & \multicolumn{1}{l}{CIF100} & \multicolumn{1}{l}{CAL101} & \multicolumn{1}{l}{CAL256} \\
\toprule
dataset\_size         & 50000                        & 50000                         & 6084                           & 30607                           \\
initial\_labeled & 1000                         & 1000                          & 1000                            & 5000                            \\
num\_cycles    & 20                           & 20                            & 5                               & 10                              \\
label\_cycle   & 1000                         & 1000                          & 1000                            & 1000                            \\
num\_epochs    & 200                          & 200                           & 200                             & 200                             \\
optimizer & SGD & SGD & SGD & SGD \\
learning\_rate         & 0.1                          & 0.1                           & 0.1    
& 0.1                             \\
momentum & 0.9 & 0.9 & 0.9 & 0.9 \\
scheduler             & step                         & step                          & step                            & step                            \\
scheduler\_step       & 160                          & 160                           & 160                             & 160                             \\
weight\_decay         & 5e-4                         & 5e-4                          & 5e-4                            & 5e-4  \\ \bottomrule
\end{tabular}
\caption{The hyperparameters used in our experimental setup.}
\label{hyperparameters}
\end{table}

\subsection{Main result: Comparisons}

\begin{figure*}[!t]
	\centering
	%\hspace{-0.0cm}
	\begin{tabular}{ccc}
	\hspace{-0.0cm}\includegraphics[scale=0.29]{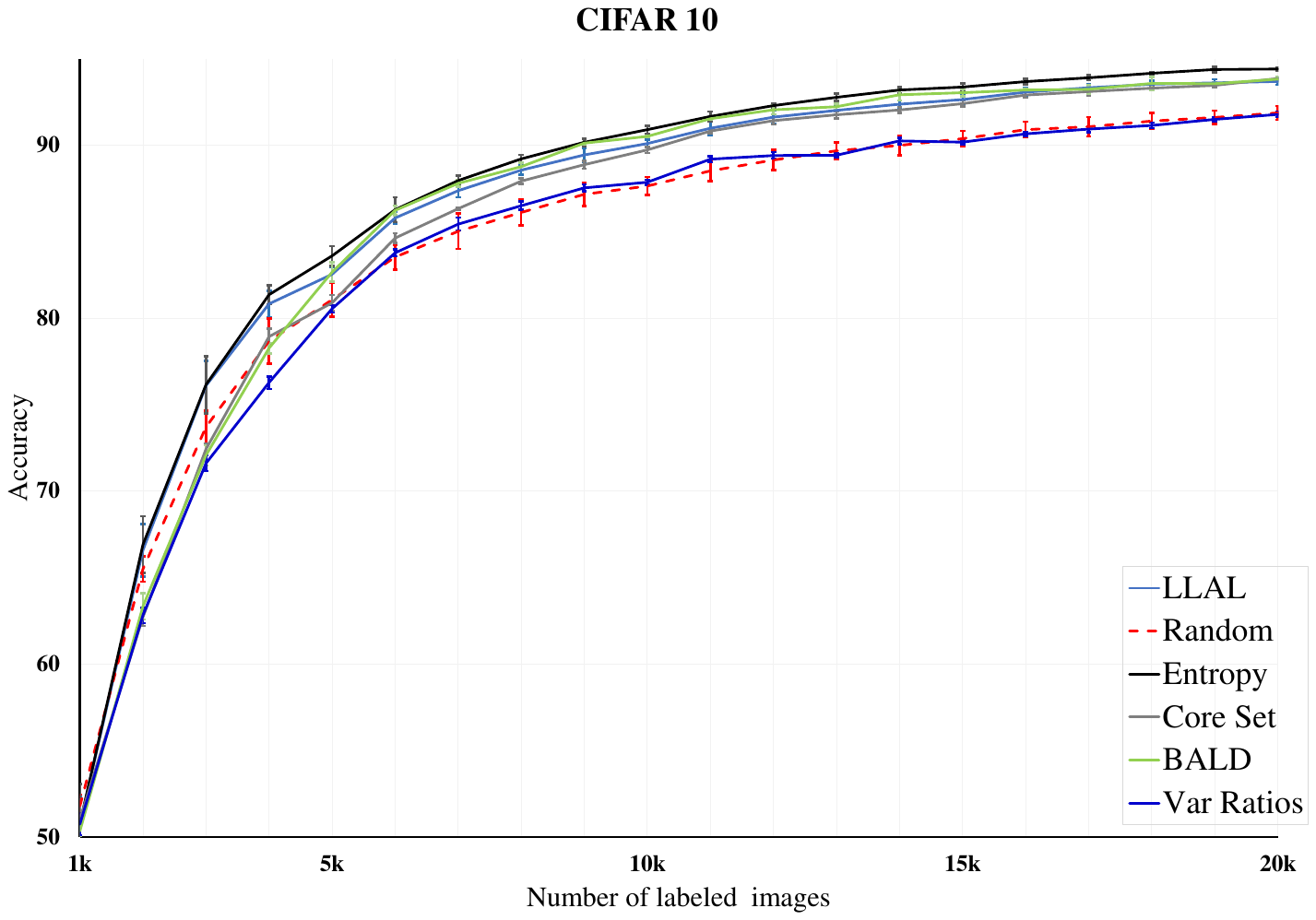}&
	\hspace{-0.2cm}\includegraphics[scale=0.29]{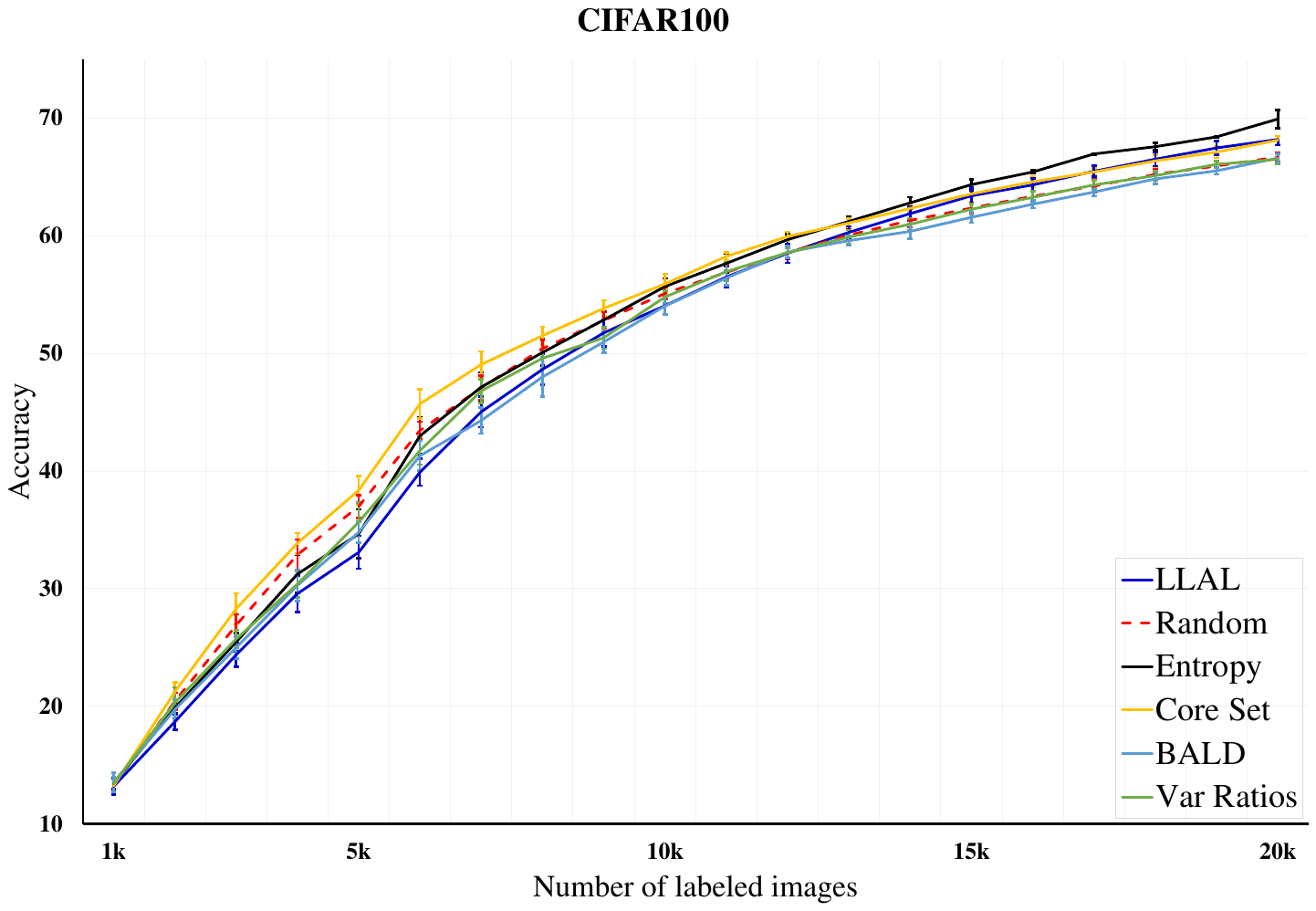}\\
    a) CIFAR-10 dataset &
    b) CIFAR-10 dataset & \\

	\hspace{-0.0cm}\includegraphics[scale=0.29]{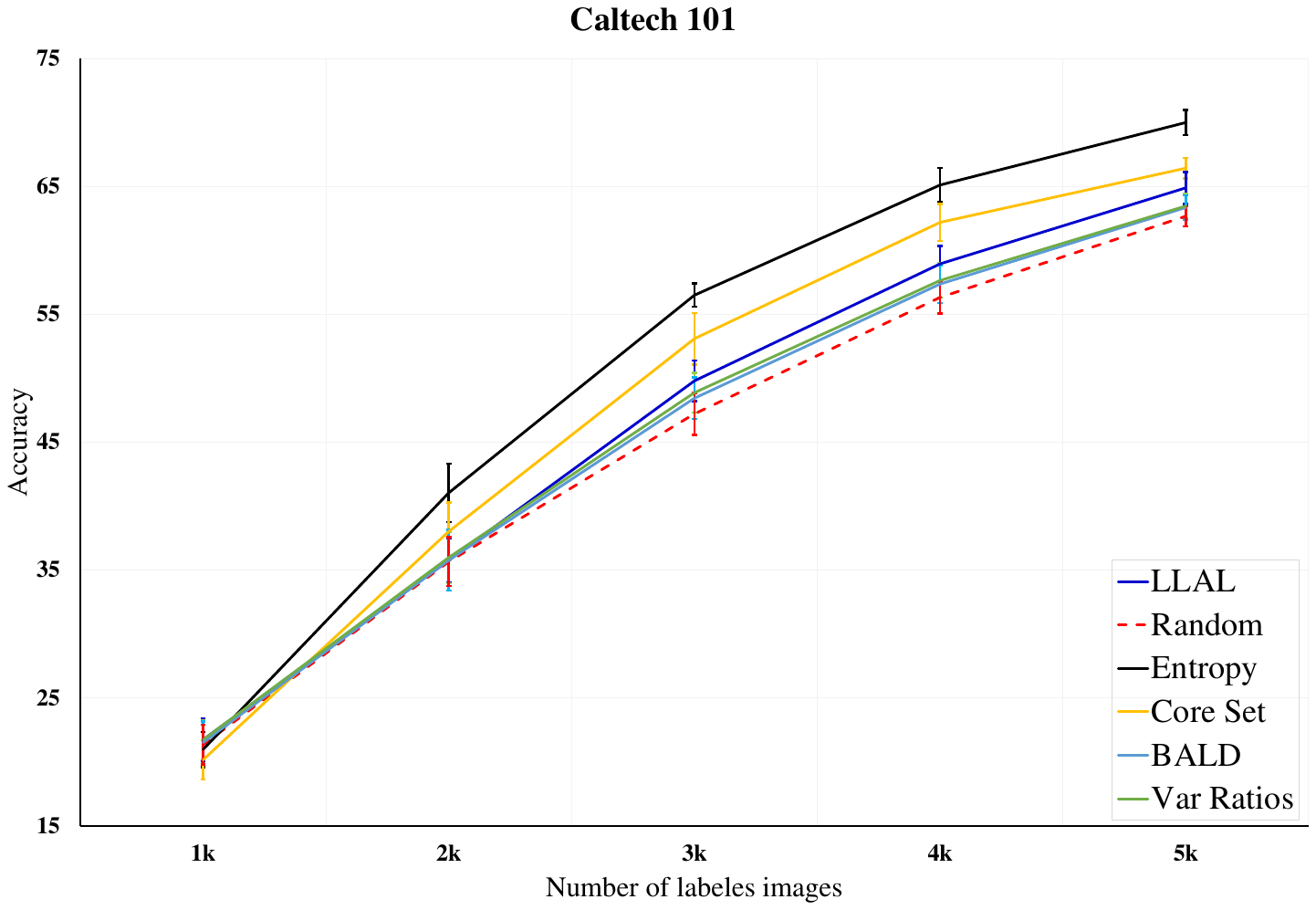}&
	\hspace{-0.2cm}\includegraphics[scale=0.29]{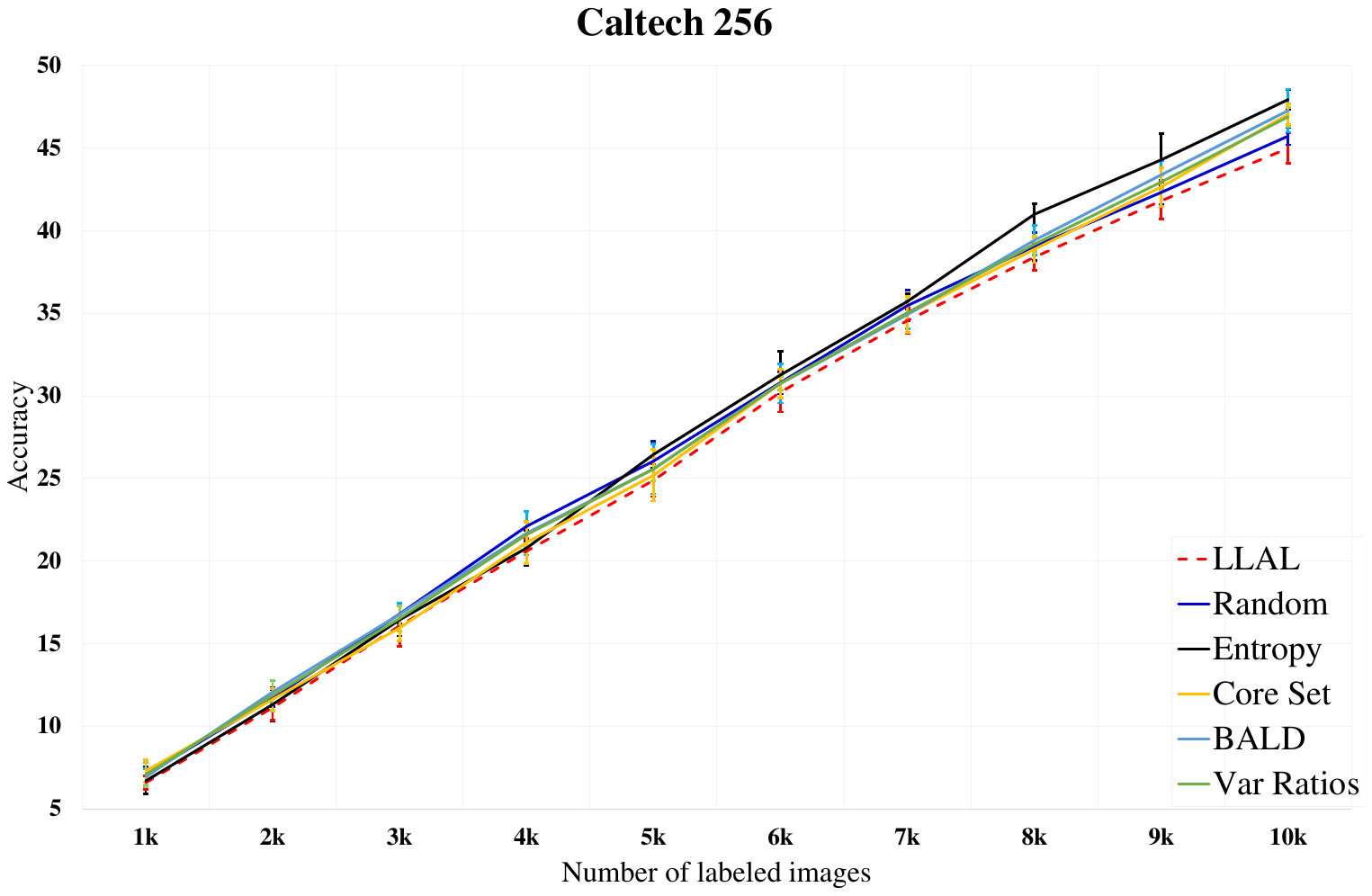}\\
    c) Caltech-101 dataset &
    d) Caltech-256 dataset & \\
	\end{tabular}
	\caption{Main results: Comparison of different active learning algorithms in a) Cifar-10, b) Cifar-100, c) Caltech-101, d) Caltech-256 datasets. We observe that the entropy (black) reaches the highest results in all datasets, with the random baseline (dotted red) typically reaching the worst results. Best viewed in high resolution when zoomed in.}
	\label{fig:main_results}
\end{figure*}

We show the results of our experiments performed in CIFAR-10 in Figure \ref{fig:main_results}a. All methods start at roughly the same point, subject to network fluctuations.
We show that immediately in the second AL cycle, the entropy and LLAL acquisition functions outperform the random baseline by $1.4$ percentage points (pp), respectively $1.1pp$. In the third cycle, both methods outperform the random baseline by over $2.5pp$. Both methods continue outperforming random by circa $3pp$ for the remaining of the training, with entropy having a slight advantage over LLAL, outperforming LLAL by around $1pp$ in the last 10 AL cycles.
Interestingly, the other methods need more AL cycles to start outperforming the random baseline. For example, the random baseline outperforms BALD and Variation Ratio in the first four AL cycles, and Core-set in the first five AL cycles.
After that, Core-set and BALD consistently outperform the random baseline, but always track behind LLAL and entropy, in the case of the latter, by up to $3pp$.
However, the Variation Ratio, after a couple of steps where it slightly outperforms the random baseline, quickly converges to the same performance and usually lags behind the entropy by $2.5-3pp$.

We see a more interesting picture when we train in the more difficult CIFAR-100, and show its results in Figure \ref{fig:main_results}b.
In the first few steps, the random baseline outperforms all the acquisition functions except the Core-set. In fact, the entropy starts clearly outperforming the random baseline only after the 12th AL step, LLAL starts outperforming the random baseline only after the 14th AL step, while the performance of BALD and Variation Ratio typically is only as good as that of the random baseline.
Core-set starts performing better than all the other methods, outperforming the random baseline by $1pp$ and entropy by $2pp$ in the first step. In the second step, it outperforms the random baseline by $1.5pp$ and entropy by almost $3pp$. It continues outperforming entropy until the 13th step, after which it consistently is outperformed by entropy, finishing in the 20th step by almost $2pp$ worse than the entropy.

We continue our experiments in the Caltech-101 dataset, which contains images with a significantly larger resolution. We show the results in Figure \ref{fig:main_results}c. In the second AL step, entropy already outperforms the random baseline by close to $5.5pp$. Core-set outperforms the random baseline by $2.5pp$ while the other three methods reach only as good results as the random baseline. Then, on the next cycle, the entropy and Core-set outperform the random baseline by around $9pp$ and respectively $6pp$. At this stage, the other three methods start outperforming the random baseline, in the case of LLAL by circa $2.5pp$ with the other two methods by around $1-1.5pp$. At the last cycle, entropy outperforms random by over $7pp$, Core-set outperforms random by around $4pp$, and LLAL outperforms random by $2.2pp$, with BALD and Variation Ratio improving over the baseline by around $1pp$.

\begin{figure*}[!t]
	\centering
	%\hspace{-0.0cm}
	\begin{tabular}{ccc}
	\hspace{-0.0cm}\includegraphics[scale=0.29]{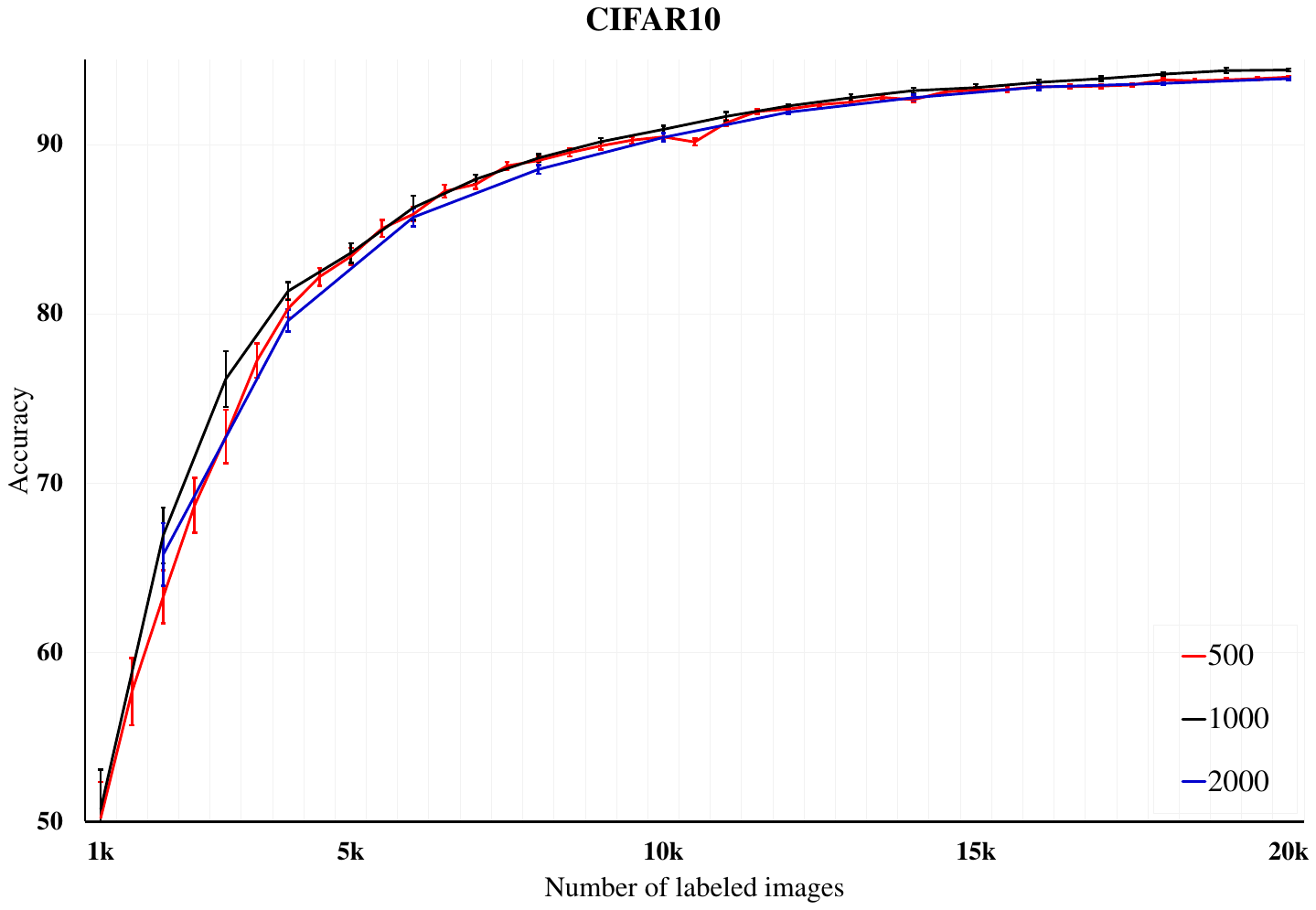}&
	\hspace{-0.2cm}\includegraphics[scale=0.29]{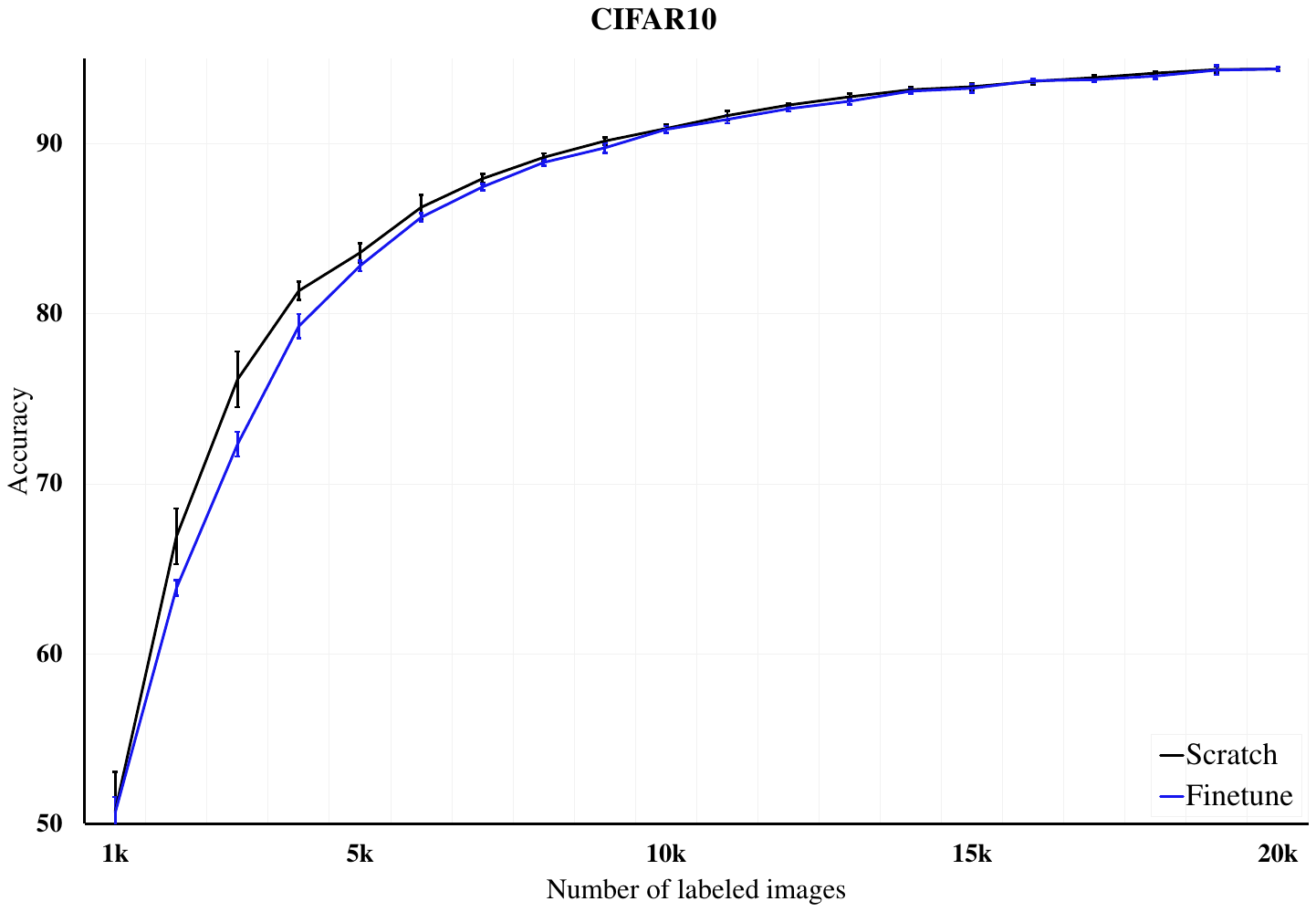}\\
    a) AL cycle labeling budget &
    b) Training vs finetuning &\\
	\end{tabular}
	\caption{a) The effect of labeling budget for active learning cycle. We observe that in general, the labeling budget should neither be too large (bias towards easy samples) nor too small (bias towards hard samples). b) The effect of training from scratch in each cycle compared to finetuning the network trained in the previous cycle. We observe that training from scratch reaches higher results, especially in the early AL cycles.}
	\label{labelingBudget_finetuning}
\end{figure*}

We now show the results in the most challenging dataset, Caltech-256 in Figure \ref{fig:main_results}d. Because of the complexity of the dataset, all the methods perform around the same in the first few AL steps. Interestingly, the random baseline outperforms all the other methods in the fourth and fifth steps. After that, the entropy takes the lead and finishes the training outperforming the random by over $4pp$. BALD, Core-set and Variation Ratio finish strongly, outperforming the random baseline by around $2.5pp$. LLAL never manages to outperform the random baseline and finishes the training getting outperformed by random by almost $1pp$.

\textbf{Recommendation 1: Entropy is all you need. } To our biggest surprise, despite the rapid research in the field of active learning, entropy is arguably still the best active learning method. While in specific scenarios, some other methods might outperform entropy, in general, entropy reaches at least competitive performances and more often than not, outperforms the other methods.
Some methods such as LLAL perform nearly as well as the entropy in the dataset they were developed. However, in more challenging datasets like CIFAR-100 and Caltech-256, they fail to show much, if any, improvement over the random baseline.
The Core-set, in general, shows the second-best performance after the entropy and usually performs highly in challenging datasets. BALD and Variation Ratio are shown to be very dataset-specific, and in general, tend to perform much worse than the entropy.
We recommend that the practitioners use the entropy acquisition function before going to more complicated solutions. More often than not, this acquisition function is the best they can get, and it tends to perform well in most if not all, settings.

\subsection{Ablation study}
We do a series of ablation studies, evaluating some of the design choices in active learning. We do the experiments in the CIFAR-10 dataset, using the best-found acquisition function, the entropy.

\subsubsection{The budget of each cycle}

Most AL research papers arbitrarily choose the number of samples labeled in every cycle. In this ablation study, we show the performance when the labeling budget for each cycle is small ($500$ samples), medium ($1000$ samples) and large ($2000$ samples). We present the results in Figure \ref{labelingBudget_finetuning}a.

As shown, until we label $4000$ samples, the medium case works significantly better than the large and especially small cases. In particular, where we have $2000$ and $3000$ labeled samples, the medium case outperforms the small case by $3.5-4pp$.
After that, the performance difference between the medium and small/large diminishes, albeit the medium case shows a slightly better performance in almost all AL cycles. At the end of the training, with $20000$ labeled samples, medium outperforms the other two cases by around half a percentage point.

\begin{figure*}[!t]
	\centering
	%\hspace{-0.0cm}
	\begin{tabular}{ccc}
	\hspace{-0.5cm}\includegraphics[scale=0.205]{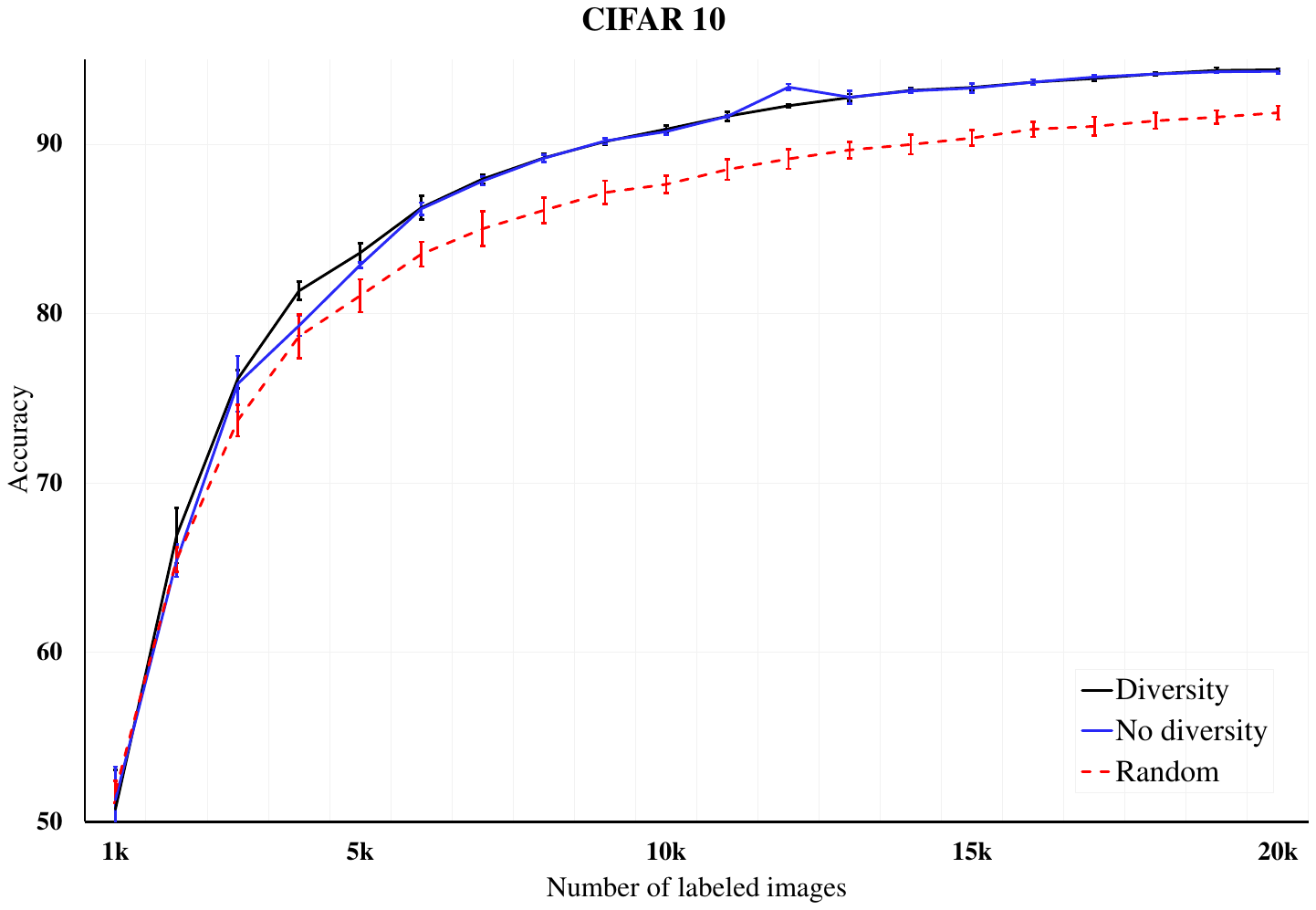} \hspace{-0.3mm}&
 	\hspace{-0.5cm}\includegraphics[scale=0.205]{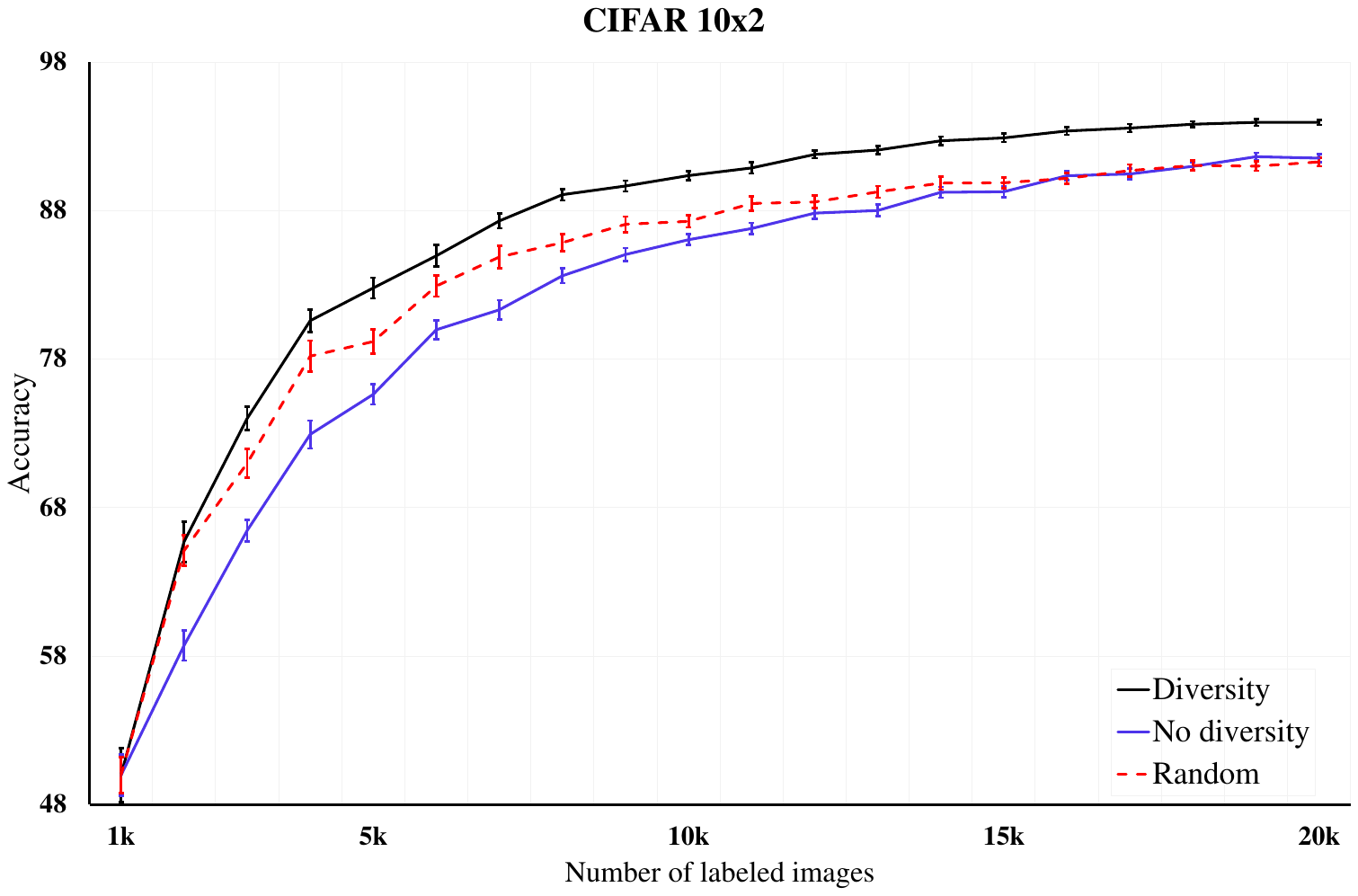} \hspace{-0.5cm}&
	\hspace{-0.5cm}\includegraphics[scale=0.205]{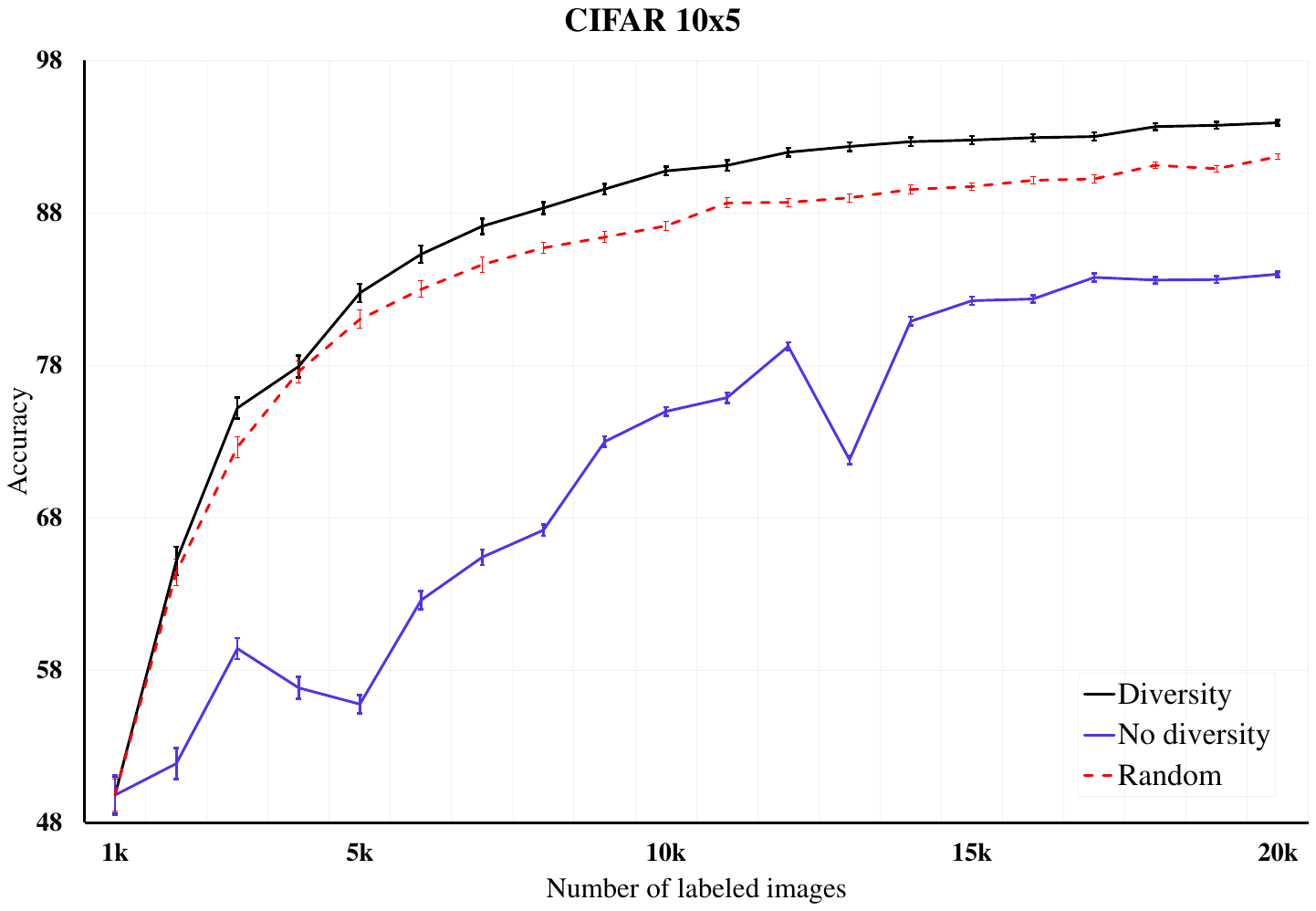} \hspace{-0.5cm}\\
    a) no repetition &
    b) 2x repetition &
    c) 5x repetition \\
	\end{tabular}
	\caption{a) Ablation study on the effect of diversity. We clearly observe that accounting for diversity matters, even if the solution is a simple heuristic. b) Ablation on diversity where each sample occurs twice in the dataset. c) Ablation on diversity where each sample occurs five times in the dataset.}
	\label{fig:diversity_od}
\end{figure*}

\textbf{Recommendation 2: Use a medium-sized budget for each AL cycle. } While using small-sized budgets might sound tempting, it does not perform very well in practice. This is because by choosing only the very hard samples, the network gets biased towards hard samples, thus, not performing well in the not-hard samples. This can be especially seen in the first few AL cycles. Similarly, using large budgets is equivalent to choosing a large number of easy, thus not very informative samples. We recommend that the practitioners use medium-sized acquisition budgets for each AL cycle.

\subsubsection{To train vs. to finetune} Some AL methods, in each AL cycle, continue training the network from the previous AL cycle. The intuition behind this is that the network has already shown some good performance, and by adding more data and continuing training, the performance of the network will improve.
On the other hand, some other works train every network from scratch. The motivation behind it is that the networks trained on a small amount of data are potentially stuck in local minima. While adding more data will improve their performance, they might still struggle to escape the local minima, thus, it is better to train them from scratch. 
We perform a full training loop, in one case training the network from \textit{scratch}, and in one training the network from the previous cycle, thus \textit{finetuning} it. We perform the experiments in CIFAR-10, using the entropy acquisition function. We present the results in Figure \ref{labelingBudget_finetuning}b.

As we can clearly see, the networks trained from scratch tend to outperform the finetuned network. In the second and third AL cycles, the networks trained from scratch outperform the finetuned networks by an average of $4pp$. The performance gain by training from scratch is lower in the next few cycles, until it completely diminishes, with both the methods performing the same.

\textbf{Recommendation 3: Train networks from scratch. } We recommend the practitioners reinitialize the network at each AL cycle. This is important, especially in the first few AL cycles where the number of data is smaller. In later cycles, it does not much difference if the network is trained from scratch or it is finetuned.

\subsubsection{The question of diversity}
Most of the methods we used in this work do not consider the diversity between the selected samples (except the Core-set). 
To consider the diversity, we adopt the heuristic of \cite{DBLP:conf/cvpr/YooK19} where we first randomly pre-select $10000$ samples to be considered, and then from them we choose the $1000$ samples to be labeled, based on the acquisition score.
This method probabilistically adds the concept of diversity to the model.
We present the results in Figure \ref{fig:diversity_od}a, showing that the method has a positive effect in the first few AL cycles. In particular, it improves the results by $1.5pp$ in the second cycle, and by a maximum of $2pp$ in the fourth AL cycle.
After the sixth cycle, we do not see any difference between the method that uses diversity and the one that does not.

We now perform another experiment where we emulate a dataset with repetition. We duplicate each sample twice (CIFAR-10x2) and five times (CIFAR-10x5).
Naturally, any method that does not consider diversity will struggle with datasets that have repetition.
This is because, for every sample that has a high acquisition score, it will also select the other identical samples to label, despite that they do not contain any extra information.
We present the results in Figure \ref{fig:diversity_od}b for Cifar-10x2, and Figure \ref{fig:diversity_od}c for Cifar-10x5
We show that the method that does not use any diversity not only suffers low results but actually performs worse than the random baseline.
Further, the higher the repetition, the worse the results of the method.
On the other hand, considering the diversity helps in not selecting identical samples, reaching significantly better results than the random baseline.

\textbf{Recommendation 4: Diversity matters. } We recommend that practitioners add the notion of diversity to their AL framework. While some methods based on constrained optimization \cite{gurobi} are computationally expensive, even simple heuristics help, especially in the early AL cycles.
Diversity is important in cases of sample repetition, and in such cases, uncertainty-based AL without diversity methods reach lower performance than the random baseline.

\begin{figure*}[!t]
	\centering
	%\hspace{-0.0cm}
	\begin{tabular}{ccc}
	\hspace{-0.5cm}\includegraphics[width=0.35\linewidth]{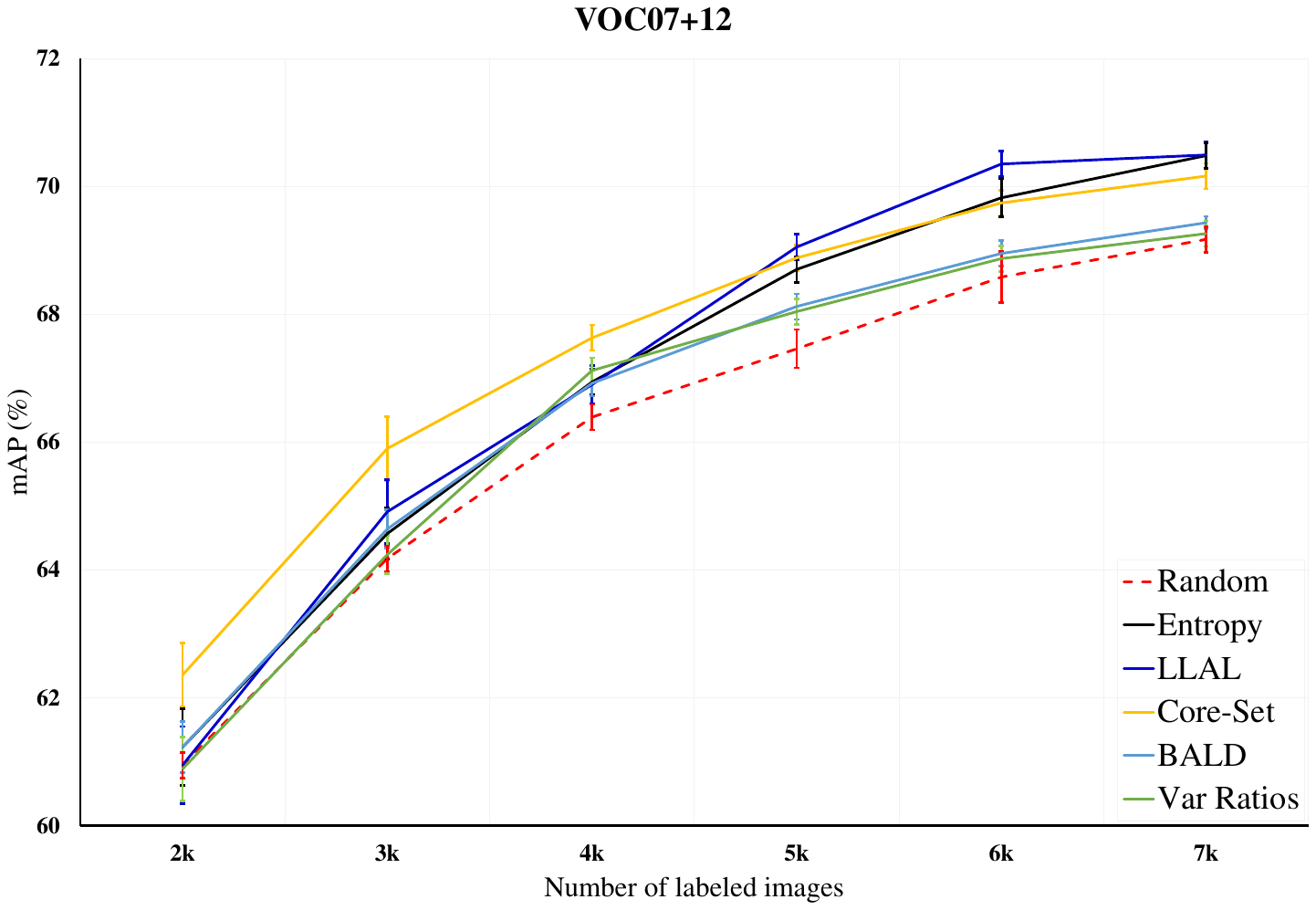} \hspace{-0.5cm}&
 	\hspace{-0.5cm}\includegraphics[width=0.35\linewidth]{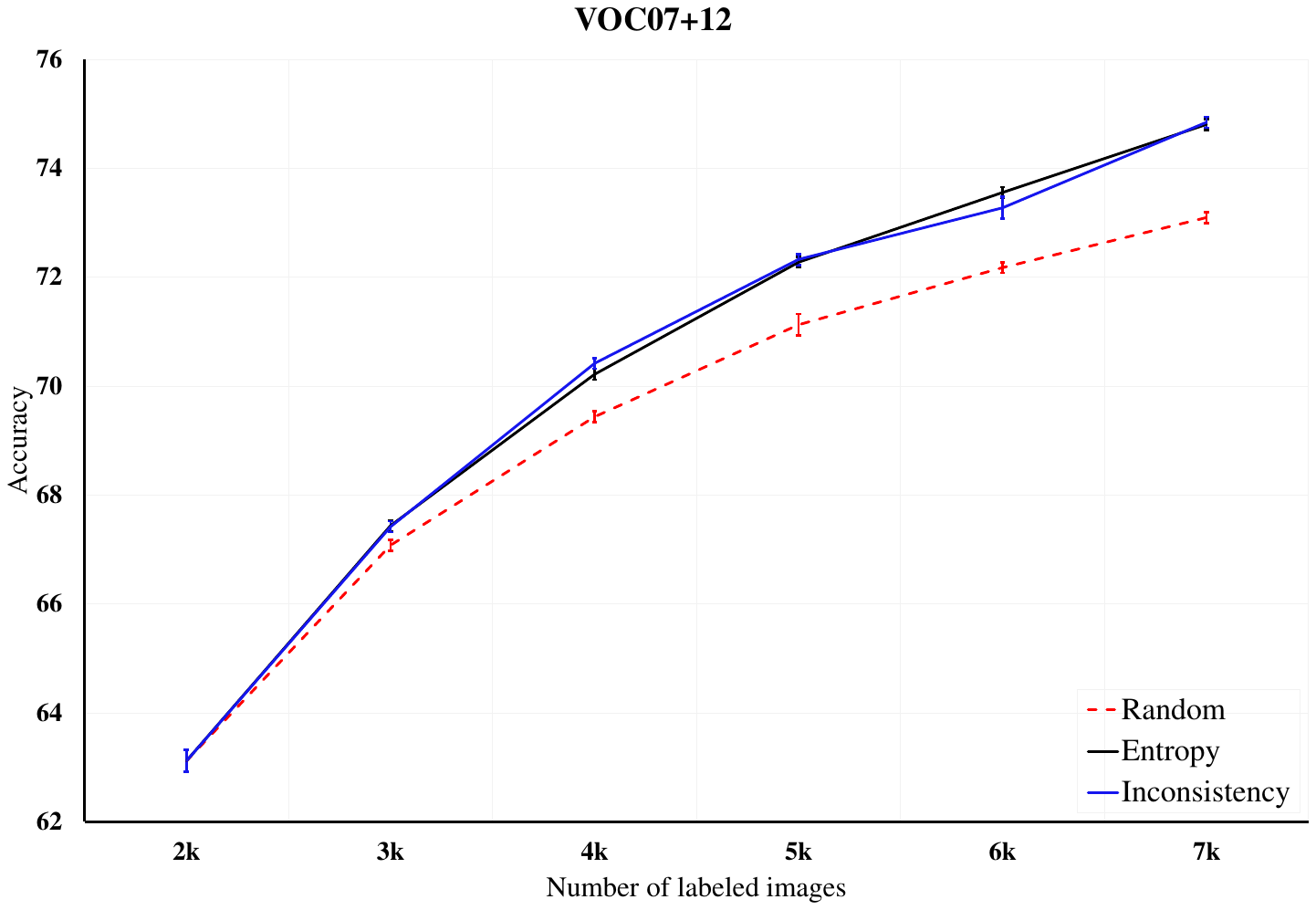} \hspace{-0.5cm}&
	\hspace{-0.5cm}\includegraphics[width=0.35\linewidth]{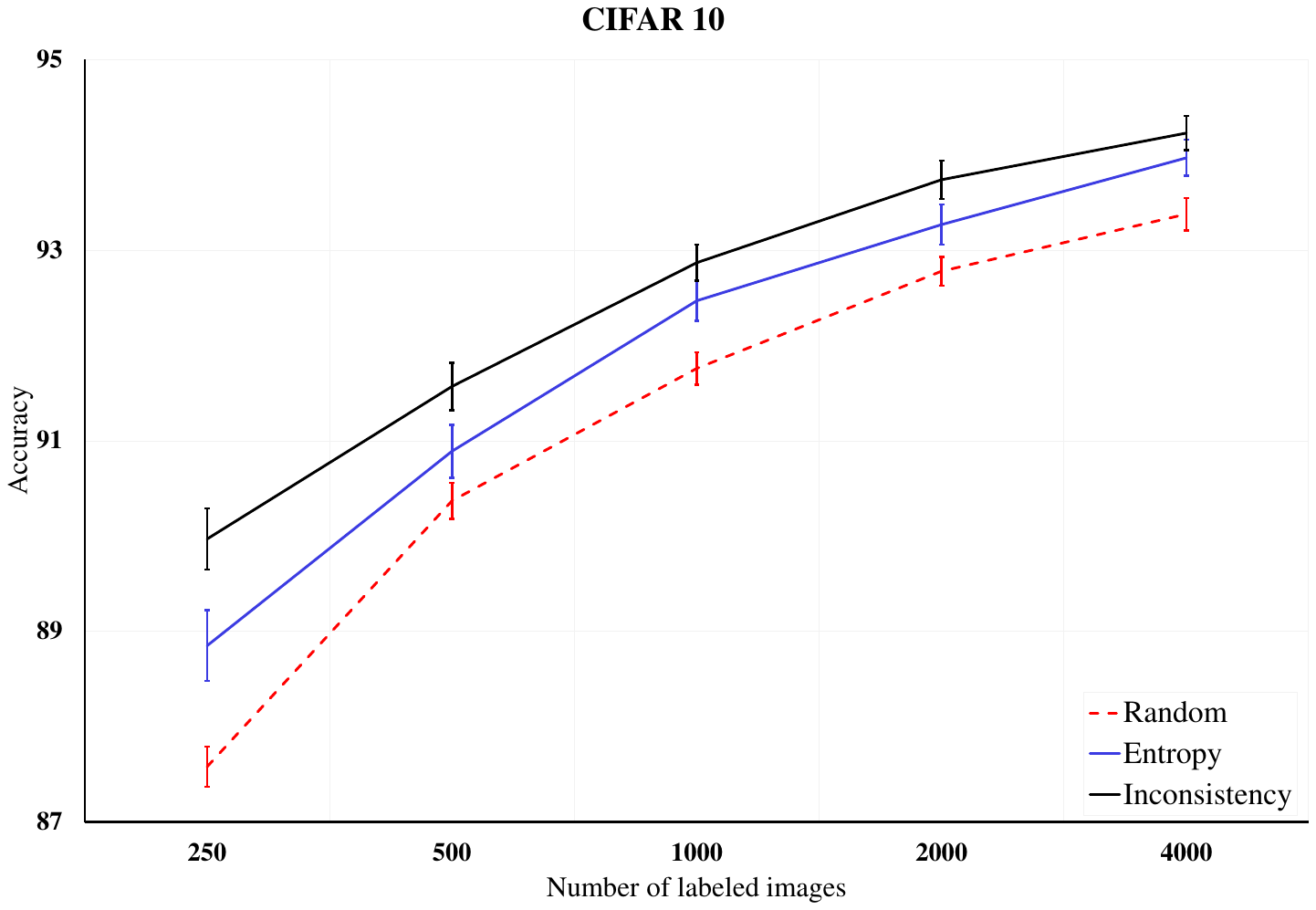} \hspace{-0.5cm}\\
    a) Detection in VOC07+12 &
    b) AL-SSL VOC07+12  &
    c) AL-SSL CIFAR-10 \\
	\end{tabular}
	\caption{a) Results in object detection. We use the PASCAL VOC07+12 dataset. b) Results of Active Learning object detection in combination with consistency-based Semi-Supervised Learning. We use the PASCAL VOC07+12 dataset. c) Results of Active Learning classification in combination with consistency-based Semi-Supervised Learning. We use the CIFAR-10 dataset.}
	\label{fig:consistency}
\end{figure*}

\subsubsection{Extension to object detection}

We now do a small study of AL methods in object detection. We extend the same six methods for object detection, not considering the AL methods that are completely tailored for object detection. We do experiments in the PASCAL VOC07+12 dataset, using SSD object detector \cite{liu2016ssd}, and the training framework of \cite{DBLP:journals/corr/abs-2103-16130}. We present our results in Figure \ref{fig:consistency}a.
In the second AL cycle, we see that all methods except Core-Set perform roughly the same. Core-Set starts very strongly, improving over the random baseline by almost $2pp$. In the third cycle, all methods outperform the random baseline by at least $0.5pp$ with Core-set reaching the best results outperforming the random baseline by more than $1pp$. In the other cycles, Core-Set, LLAL and entropy reach around the same results, all of them ending the training by at least $1pp$ better than the random baseline. However, BALD and Variational Ratio just slightly improved over the random baseline.

\textbf{Recommendation 5: Use AL for object detection. } However, do not expect the performance improvement to be as big as in classification. From the evaluated methods entropy and Core-set seem to perform the best. From other studies, methods tailored for object detection tend to outperform our baselines \cite{DBLP:journals/corr/abs-2103-16130,DBLP:conf/iccv/AghdamGLW19,DBLP:conf/accv/KaoLS018,DBLP:conf/bmvc/RoyUN18,haussmann2020scalable,DBLP:conf/cvpr/YuanWFLXJY21}, so we recommend considering them in addition to mentioned methods.

\subsubsection{Consistency helps active learning}

We combine Active Learning with consistency-based Semi-Supervised Learning (AL-SSL) and see their combined effects.
For classification, we follow previous work and use \cite{DBLP:conf/nips/BerthelotCGPOR19} as the SSL algorithm of choice.
We use an adaption of the method \cite{DBLP:conf/nips/JeongLKK19} for object detection.
We show the results of the SSL when done without active learning (random baseline), with entropy acquisition function and with inconsistency acquisition function \cite{DBLP:conf/eccv/GaoZYADP20}.
For the inconsistency acquisition function, we use the consistency loss as done in \cite{al-ssl-od}.
We provide all the implementation details in the supplementary material.
We present the classification results in Figure \ref{fig:consistency}c. We see that the results of all methods are significantly higher than those in the previous experiments.
This is because of the effect of the semi-supervised learning, that considerably improves the results. 
However, we can also see the effect of Active Learning, which comes at the top of Semi-Supervised Learning.
We see that both the entropy and consistency considerably improve over the random baseline, with inconsistency-based AL reaching the best results
Similarly, we present the results of object detection in Figure \ref{fig:consistency}b.
We observe that while in the second AL cycle, there is just a small improvement between the AL-SSL methods, and the SSL method with random acquisition score, in the later cycles, we see a significant improvement. For example, entropy outperforms random by more than $1pp$ in the last three AL cycles, with a peak of $1.7pp$ performance improvement in the last cycle. 
We also observe that the performance of the entropy acquisition function and the inconsistency AL are very close to each other, with the methods performing as well as each other.

\textbf{Recommendation 6: Combine SSL with AL.} The results will significantly improve, and the two methodologies improve each other. 
\vspace{2cm}
\section{Conclusion}
\vspace{-0.2cm}
In this study, we performed a fair empirical study of various Active Learning methods and ablated some of the most important parts of the models.
We performed the study in a controlled setting, where we took care to minimize the possible fluctuations of the network training, hyperparameters and initial labeling set.
%
%We performed our study in 4 standard computer vision datasets, using 6 famous acquisition functions.
%
Our most interesting finding is surprising: in general, no method tends to outperform entropy, the simplest acquisition function.
We do ablations in the budget for each AL cycle, training or finetuning the network, and handling the diversity in the dataset, giving recommendations for each of them.
We end our study by showing experiments that combined AL with SSL, and extensions in object detection.
\section*{Broader Impact}

Labeling data is one of the biggest pitfalls in machine learning cycles. It is an expensive process and subject to human errors. 
In some settings, such as autonomous driving, the majority of data is redundant, thus labeling more of it, does not come with a performance improvement.
AL is one of the most promising approaches that helps practitioners in choosing to label the right data, which when fed to a network, give the best improvement.
Despite many AL methods claiming to be the SOTA in the field, there are many questions with respect to the experimental setup, and thus the lessons learned from them.
In this work, we shed some light on the world of AL, devising a fair setup, and training multiple AL methods in different datasets.
We trained over $3,000$ networks, showing the best-performing AL method, and learning in the process some insights that we hope can help other researchers.
Similar studies in SSL \cite{DBLP:conf/nips/OliverORCG18} and metric learning \cite{DBLP:conf/eccv/MusgraveBL20} gave a massive boost to the research in their field. 
We hope our study can help practitioners choose the right acquisition function and engineering practices so that not everyone needs to reinvent the wheel.

%%%%%%%%% REFERENCES
{\small
\bibliographystyle{ieee_fullname}
\bibliography{egbib}
}

\clearpage

%%%%%%%%% ABSTRACT

In this supplementary material, we complement the results of the main paper. We give extra information about the consistency sections (Section 4.3.5 in the main paper), and we provide the mean and standard deviation for every experiment. We perform each experiment in a single NVIDIA V100 GPU.

\subsection{Consistency helps active learning}
\subsection{Image classification}
We combine Active Learning with consistency-based Semi-Supervised Learning (AL-SSL) and see their combined effects.
Because the results of SSL are already quite high with $1k-2k$ labeled images, we perform the experiments starting with $250$ images, and double the number of labels in each cycle. 
For classification, we follow previous work and use MixMatch \cite{DBLP:conf/nips/BerthelotCGPOR19,DBLP:conf/eccv/GaoZYADP20} as the SSL algorithm of choice.
We train the network for $1024$ epochs, we do not make any changes to the algorithm, training procedure, or the backbone.
During training, MixMatch makes two different augmentations in an image, and computes the loss function as the distance between the network's prediction
We use the same loss function as inconsistency acquisition score.

\subsubsection{Object detection}
We present the results of the SSL in object detection.
As SSL method, we use that of \cite{DBLP:conf/nips/JeongLKK19}. The method feed to the network an image, and its augmented version (for example, by performing a horizontal clip), and then computes as loss function the symmetric KL divergence between the predictions of the network for both views of the image.
Similar to \cite{al-ssl-od}, we use the same loss function as acquisition function.
We train the model for $120,000$ iterations using SGD with momentum. and we do not make any changes to the algorithm or the training procedure.

\subsection{Detailed results}
In the paper, we provide plots for the main experiments
due to the limited space. In Tables \ref{supp:cifar-10}, \ref{supp:cifar-100}, \ref{supp:caltech101}, \ref{supp:caltech256}, \ref{supp:budget}, \ref{supp:train_finetune}, \ref{supp:diversity}, \ref{supp:diversityx2}, \ref{supp:diversityx5}, \ref{supp:voc07+12}, \ref{supp:cons_voc07+12}, \ref{supp:cons} we summarize the exact numbers corresponding to Figures 1a,
1b, 1c, 1d, 2a, 2b, 3a, 3b, 3c, 4a, 4b and 4c of the main paper. We provide the
mean and the standard deviation for each method and AL
cycle. Each experiment has been run five times.

\begin{table*}[!thb]
\centering
\begin{tabular}{l|llllll}
\toprule
\# labeled  & Random & LLAL                                              & Entropy                     & Core Set                    & BALD                        & Var Ratio                  \\ \hline
1k     & \textbf{51.78$\pm$0.6}             & 50.74$\pm$1.5 & 50.74$\pm$2.3 & 51.01$\pm$1.1 & 50.34$\pm$0.3 & 50.74$\pm$0.6 \\
2k             & 65.48$\pm$0.7    & 66.58$\pm$1.5  & \textbf{66.90$\pm$1.6} & 62.71$\pm$0.5 & 63.33$\pm$0.8 & 62.82$\pm$0.5 \\
3k        & 73.69$\pm$0.9         & 76.09$\pm$1.4  & \textbf{76.14$\pm$1.6} & 72.42$\pm$0.3 & 72.06$\pm$0.6 & 71.60$\pm$0.5 \\
4k          & 78.66$\pm$1.3       & 80.81$\pm$0.8  & \textbf{81.34$\pm$0.5} & 78.92$\pm$0.5 & 78.26$\pm$0.3 & 76.27$\pm$0.4 \\
5k         & 81.05$\pm$1.0        & 82.52$\pm$0.4  & \textbf{83.59$\pm$0.6} & 80.88$\pm$0.5 & 82.65$\pm$0.6 & 80.53$\pm$0.2 \\
6k        & 83.52$\pm$0.7         & 85.78$\pm$0.3  & \textbf{86.26$\pm$0.7} & 84.62$\pm$0.3 & 86.23$\pm$0.3 & 83.77$\pm$0.2 \\
7k         & 85.01$\pm$1.0        & 87.36$\pm$0.4  & \textbf{87.95$\pm$0.3} & 86.32$\pm$0.1 & 87.79$\pm$0.2 & 85.43$\pm$0.4 \\
8k         & 86.10$\pm$0.7        & 88.55$\pm$0.3  & \textbf{89.19$\pm$0.2} & 87.92$\pm$0.2 & 88.75$\pm$0.2 & 86.49$\pm$0.2 \\
9k         & 87.15$\pm$0.7        & 89.42$\pm$0.4  & \textbf{90.16$\pm$0.2} & 88.87$\pm$0.2 & 90.11$\pm$0.1 & 87.53$\pm$0.2 \\
10k        & 87.63$\pm$0.5        & 90.08$\pm$0.2  & \textbf{90.89$\pm$0.2} & 89.72$\pm$0.2 & 90.50$\pm$0.1 & 87.85$\pm$0.1 \\
11k       & 88.51$\pm$0.6         & 90.97$\pm$0.4  & \textbf{91.65$\pm$0.3} & 90.81$\pm$0.1 & 91.53$\pm$0.1 & 89.19$\pm$0.2 \\
12k        & 89.13$\pm$0.6        & 91.61$\pm$0.4  & \textbf{92.27$\pm$0.1} & 91.41$\pm$0.2 & 92.04$\pm$0.3 & 89.40$\pm$0.2 \\
13k        & 89.66$\pm$0.5        & 92.00$\pm$0.2  & \textbf{92.76$\pm$0.2} & 91.75$\pm$0.2 & 92.21$\pm$0.2 & 89.41$\pm$0.1 \\
14k         & 89.98$\pm$0.6       & 92.36$\pm$0.2  & \textbf{93.18$\pm$0.2} & 92.03$\pm$0.2 & 92.91$\pm$0.3 & 90.24$\pm$0.2 \\
15k         & 90.37$\pm$0.5       & 92.63$\pm$0.2  & \textbf{93.35$\pm$0.2} & 92.40$\pm$0.2 & 93.03$\pm$0.2 & 90.17$\pm$0.1 \\
16k          & 90.89$\pm$0.4      & 93.06$\pm$0.2  & \textbf{93.67$\pm$0.2} & 92.89$\pm$0.2 & 93.18$\pm$0.1 & 90.65$\pm$0.1 \\
17k         & 91.06$\pm$0.6       & 93.31$\pm$0.2  & \textbf{93.89$\pm$0.1} & 93.09$\pm$0.2 & 93.22$\pm$0.2 & 90.92$\pm$0.2 \\
18k          & 91.39$\pm$0.5      & 93.52$\pm$0.2  & \textbf{94.15$\pm$0.1} & 93.29$\pm$0.1 & 93.56$\pm$0.4 & 91.14$\pm$0.2 \\
19k          & 91.60$\pm$0.4      & 93.60$\pm$0.2  & \textbf{94.36$\pm$0.2} & 93.45$\pm$0.1 & 93.55$\pm$0.1 & 91.48$\pm$0.1 \\
20k       & 91.86$\pm$0.4         & 93.67$\pm$0.2  & \textbf{94.40$\pm$0.1} & 93.85$\pm$0.1 & 93.81$\pm$0.2 & 91.77$\pm$0.1 \\ \bottomrule
\end{tabular}
\caption{CIFAR-10 detailed results. This table corresponds to results of Figure 1a in the main paper.}
\label{supp:cifar-10}
\end{table*}

\begin{table*}[!thb]
\centering
\begin{tabular}{l|llllll}
\toprule
\# labeled  & Random & LLAL                                              & Entropy                     & Core Set                    & BALD                        & Var Ratio                  \\ \hline
1k                 & 13.28$\pm$0.7 & 13.19$\pm$0.7 & 13.42$\pm$0.5 & 13.24$\pm$0.7 & \textbf{13.56}$\pm$\textbf{0.8} & 13.39$\pm$0.6 \\
2k                 & 20.43$\pm$0.7 & 18.69$\pm$0.7 & 19.87$\pm$0.4 & \textbf{21.24$\pm$0.8} & 19.73$\pm$0.9 & 20.32$\pm$1.3 \\
3k                 & 26.89$\pm$1.0 & 24.37$\pm$0.9 & 25.47$\pm$0.8 & \textbf{28.28$\pm$1.3} & 25.00$\pm$0.9 & 25.72$\pm$0.8 \\
4k                 & 32.88$\pm$1.6 & 29.57$\pm$1.3 & 31.24$\pm$1.6 & \textbf{33.87$\pm$0.9} & 30.28$\pm$1.4 & 30.38$\pm$1.1 \\
5k                 & 36.99$\pm$1.4 & 33.09$\pm$1.0 & 34.66$\pm$2.1 & \textbf{38.36$\pm$1.2} & 34.79$\pm$0.9 & 35.66$\pm$1.7 \\
6k                 & 43.45$\pm$1.1 & 39.90$\pm$0.7 & 43.00$\pm$1.6 & \textbf{45.70$\pm$1.3} & 41.30$\pm$1.3 & 41.73$\pm$1.2 \\
7k                 & 47.08$\pm$1.3 & 45.03$\pm$1.0 & 47.13$\pm$1.2 & \textbf{49.06$\pm$1.1} & 44.29$\pm$1.1 & 46.81$\pm$1.0 \\
8k                 & 50.41$\pm$1.3 & 48.65$\pm$0.7 & 50.08$\pm$1.1 & \textbf{51.51$\pm$0.8} & 48.01$\pm$1.7 & 49.58$\pm$0.9 \\
9k                 & 52.80$\pm$1.1 & 51.74$\pm$0.7 & 52.85$\pm$0.7 & \textbf{53.83$\pm$0.7} & 50.97$\pm$0.9 & 51.32$\pm$0.9 \\
10k                & 55.09$\pm$0.8 & 54.05$\pm$0.5 & 55.68$\pm$0.7 & \textbf{55.90$\pm$0.9} & 54.03$\pm$0.7 & 54.79$\pm$0.5 \\
11k                & 56.89$\pm$0.9 & 56.49$\pm$0.6 & 57.66$\pm$0.8 & \textbf{58.25$\pm$0.4} & 56.41$\pm$0.6 & 56.95$\pm$0.8 \\
12k                & 58.58$\pm$0.8 & 58.51$\pm$0.6 & 59.67$\pm$0.5 & \textbf{59.93$\pm$0.4} & 58.61$\pm$0.5 & 58.56$\pm$0.4 \\
13k                & 60.05$\pm$0.5 & 60.29$\pm$0.5 & \textbf{61.21$\pm$0.4} & 61.09$\pm$0.4 & 59.58$\pm$0.4 & 59.89$\pm$0.6 \\
14k                & 61.30$\pm$0.6 & 61.88$\pm$0.6 & \textbf{62.80$\pm$0.5} & 62.33$\pm$0.3 & 60.38$\pm$0.6 & 60.97$\pm$0.5 \\
15k                & 62.35$\pm$0.7 & 63.39$\pm$0.5 & \textbf{64.35$\pm$0.5} & 63.56$\pm$0.7 & 61.56$\pm$0.5 & 62.25$\pm$0.4 \\
16k                & 63.35$\pm$0.6 & 64.32$\pm$0.4 & \textbf{65.42$\pm$0.1} & 64.58$\pm$0.4 & 62.67$\pm$0.3 & 63.26$\pm$0.5 \\
17k                & 64.22$\pm$0.5 & 65.48$\pm$0.6 & \textbf{66.96$\pm$0.0} & 65.42$\pm$0.3 & 63.72$\pm$0.3 & 64.32$\pm$0.4 \\
18k                & 65.21$\pm$0.6 & 66.51$\pm$0.5 & \textbf{67.58$\pm$0.3} & 66.36$\pm$0.5 & 64.83$\pm$0.4 & 65.11$\pm$0.4 \\
19k                & 65.92$\pm$0.6 & 67.46$\pm$0.4 & \textbf{68.41$\pm$0.0} & 67.11$\pm$0.5 & 65.52$\pm$0.3 & 66.07$\pm$0.3 \\
20k                & 66.69$\pm$0.5 & 68.18$\pm$0.4 & \textbf{69.92$\pm$0.8} & 68.12$\pm$0.4 & 66.60$\pm$0.4 & 66.50$\pm$0.4\\ \bottomrule
\end{tabular}
\caption{CIFAR-100 detailed results. This table corresponds to results of Figure 1b in the main paper.}
\label{supp:cifar-100}
\end{table*}
\begin{table*}[!thb]
\centering
\begin{tabular}{l|llllll}
\toprule
\# labeled  & Random & LLAL                                              & Entropy                     & Core Set                    & BALD                        & Var Ratio                  \\ \hline
1k      & 21.35$\pm$1.6 & \textbf{21.75$\pm$1.7} & 20.98$\pm$1.4 & 20.17$\pm$1.5 & 21.52$\pm$1.6 & 21.74$\pm$1.5 \\
2k      & 35.66$\pm$1.9 & 35.76$\pm$1.7 & \textbf{41.04$\pm$2.3} & 38.02$\pm$2.3 & 35.75$\pm$2.4 & 35.98$\pm$2.0 \\
3k      & 47.21$\pm$1.6 & 49.79$\pm$1.6 & \textbf{56.49$\pm$0.9} & 53.09$\pm$2.0 & 48.43$\pm$1.6 & 48.87$\pm$1.5 \\
4k      & 56.32$\pm$1.3 & 58.95$\pm$1.4 & \textbf{65.10$\pm$1.3} & 62.18$\pm$1.4 & 57.36$\pm$1.5 & 57.64$\pm$1.3 \\
5k      & 62.66$\pm$0.8 & 64.88$\pm$1.2 & \textbf{69.99$\pm$1.0} & 66.42$\pm$0.8 & 63.34$\pm$0.9 & 63.46$\pm$0.9\\ \bottomrule
\end{tabular}
\caption{Caltech101 detailed results. This table corresponds to results of Figure 1c in the main paper.}
\label{supp:caltech101}
\end{table*}
\begin{table*}[!thb]
\centering
\begin{tabular}{l|llllll}
\toprule
\# labeled  & Random & LLAL                                              & Entropy                     & Core Set                    & BALD                        & Var Ratio                  \\ \hline
1k      & 7.06$\pm$0.5  & 6.58$\pm$0.4  & 6.69$\pm$0.8  & \textbf{7.29$\pm$0.7}  & 6.93$\pm$0.5  & 7.06$\pm$0.7  \\
2k      & 11.70$\pm$0.5 & 11.12$\pm$0.8 & 11.31$\pm$1.0 & 11.62$\pm$0.7 & 12.07$\pm$0.7 & \textbf{11.86$\pm$0.9} \\
3k      & \textbf{16.80$\pm$0.6} & 16.07$\pm$1.3 & 16.44$\pm$1.0 & 15.95$\pm$0.8 & 16.78$\pm$0.7 & 16.53$\pm$0.8 \\
4k      & 22.11$\pm$0.9 & 20.60$\pm$0.8 & 20.79$\pm$1.1 & 21.12$\pm$1.3 & \textbf{21.68$\pm$1.3} & 21.60$\pm$0.8 \\
5k      & 26.05$\pm$1.2 & 24.89$\pm$1.0 & \textbf{26.45$\pm$0.8} & 25.19$\pm$1.6 & 25.58$\pm$1.5 & 25.56$\pm$0.7 \\
6k      & 30.80$\pm$0.7 & 30.23$\pm$1.2 & \textbf{31.27$\pm$1.4} & 30.75$\pm$0.9 & 30.76$\pm$1.2 & 30.76$\pm$0.4 \\
7k      & 35.47$\pm$0.9 & 34.56$\pm$0.8 & \textbf{35.71$\pm$0.5} & 34.94$\pm$1.1 & 34.92$\pm$0.9 & 35.02$\pm$0.4 \\
8k      & 39.04$\pm$0.8 & 38.40$\pm$0.8 & \textbf{40.99$\pm$0.7} & 38.88$\pm$0.8 & 39.42$\pm$0.9 & 39.18$\pm$0.4 \\
9k      & 42.32$\pm$0.7 & 41.81$\pm$1.1 & \textbf{44.30$\pm$1.6} & 42.65$\pm$1.2 & 43.39$\pm$0.9 & 42.95$\pm$0.4 \\
    10k     & 45.73$\pm$0.5 & 45.00$\pm$0.9 & \textbf{47.95$\pm$0.6} & 47.05$\pm$0.6 & 47.28$\pm$1.3 & 46.90$\pm$0.6\\ \bottomrule
\end{tabular}
\caption{Caltech256 detailed results. This table corresponds to results of Figure 1d in the main paper.}
\label{supp:caltech256}
\end{table*}
\begin{table*}[!thb]
\centering
\begin{tabular}{l|llllll}
\toprule
\#labeled & 500        & 1000       & 2000       \\
\hline
500     & 41.50$\pm$3.3 &            &            \\
1k      & 50.22$\pm$2.1 & \textbf{50.74$\pm$2.3} &            \\
1.5k    & 57.67$\pm$1.9 &            &            \\
2k      & 63.28$\pm$1.5 & \textbf{66.90$\pm$1.6} & 65.78$\pm$1.8 \\
2.5k    & 68.68$\pm$1.6 &            &            \\
3k      & 72.75$\pm$1.5 & \textbf{76.14$\pm$1.6} &            \\
3.5k    & 77.24$\pm$1.0 &            &            \\
4k      & 80.32$\pm$0.5 & \textbf{81.3$\pm$0.5}  & 79.60$\pm$0.6 \\
4.5k    & 82.19$\pm$0.5 &            &            \\
5k      & 83.39$\pm$0.5 & \textbf{83.59$\pm$0.5} &            \\
5.5k    & 85.04$\pm$0.5 &            &            \\
6k      & 85.88$\pm$0.4 & \textbf{86.26$\pm$0.7} & 85.70$\pm$0.5 \\
6.5k    & 87.24$\pm$0.3 &            &            \\
7k      & 87.65$\pm$0.2 & \textbf{87.95$\pm$0.2} &            \\
7.5k    & 88.74$\pm$0.2 &            &            \\
8k      & 89.05$\pm$0.2 & \textbf{89.19$\pm$0.2} & 88.53$\pm$0.2 \\
8.5k    & 89.52$\pm$0.2 &            &            \\
9k      & 89.92$\pm$0.2 & \textbf{90.16$\pm$0.2} &            \\
9.5k    & \textbf{90.26$\pm$0.2} &            &            \\
10k     & 90.45$\pm$0.1 & \textbf{90.89$\pm$0.2} & 90.42$\pm$0.2 \\
10.5k   & 90.15$\pm$0.1 &            &            \\
11k     & 91.27$\pm$0.1 & \textbf{91.65$\pm$0.2} &            \\
11.5k   & 91.95$\pm$0.1 &            &            \\
12k     & 92.09$\pm$0.1 & \textbf{92.27$\pm$0.1} & 91.91$\pm$0.1 \\
12.5k   & 92.35$\pm$0.1 &            &            \\
13k     & 92.50$\pm$0.1 & \textbf{92.76$\pm$0.2} &            \\
13.5k   & 92.78$\pm$0.1 &            &            \\
14k     & 92.65$\pm$0.1 & \textbf{93.18$\pm$0.1} & 92.76$\pm$0.1 \\
14.5k   & 93.11$\pm$0.1 &            &            \\
15k     & 93.22$\pm$0.1 & \textbf{93.35$\pm$0.2} &            \\
15.5k   & 93.25$\pm$0.1 &            &            \\
16k     & 93.42$\pm$0.1 & \textbf{93.67$\pm$0.1} & 93.39$\pm$0.1 \\
16.5k   & 93.42$\pm$0.1 &            &            \\
17k     & 93.45$\pm$0.1 & \textbf{93.89$\pm$0.1} &            \\
17.5k   & 93.52$\pm$0.0 &            &            \\
18k     & 93.82$\pm$0.0 & \textbf{94.15$\pm$0.0} & 93.61$\pm$0.0 \\
18.5k   & 93.76$\pm$0.0 &            &            \\
19k     & 93.82$\pm$0.0 & \textbf{94.36$\pm$0.1} &            \\
19.5k   & 93.89$\pm$0.0 &            &            \\
20k     & 93.98$\pm$0.0 & \textbf{94.40$\pm$0.0} & 93.89$\pm$0.0\\ \bottomrule
\end{tabular}
\caption{CIFAR-10 detailed results in the budget size. This table corresponds to results of Figure 2a in the main paper.}
\label{supp:budget}
\end{table*}
\begin{table*}[!thb]
\centering
\begin{tabular}{l|llllll}
\toprule
\#labeled & Scratch    & Finetune   \\
\hline
1k        & \textbf{50.74$\pm$0.3} & 50.66$\pm$0.9 \\
2k        & \textbf{66.90$\pm$1.6} & 63.87$\pm$0.4 \\
3k        & \textbf{76.14$\pm$1.6} & 72.33$\pm$0.7 \\
4k        & \textbf{81.34$\pm$0.5} & 79.27$\pm$0.7 \\
5k        & \textbf{83.59$\pm$0.5} & 82.82$\pm$0.3 \\
6k        & \textbf{86.26$\pm$0.7} & 85.68$\pm$0.2 \\
7k        & \textbf{87.95$\pm$0.2} & 87.46$\pm$0.1 \\
8k        & \textbf{89.19$\pm$0.2} & 88.90$\pm$0.2 \\
9k        & \textbf{90.16$\pm$0.2} & 89.76$\pm$0.3 \\
10k       & \textbf{90.89$\pm$0.2} & 90.85$\pm$0.1 \\
11k       & \textbf{91.65$\pm$0.2} & 91.43$\pm$0.2 \\
12k       & \textbf{92.27$\pm$0.1} & 92.06$\pm$0.1 \\
13k       & \textbf{92.76$\pm$0.2} & 92.5$\pm$0.1  \\
14k       & \textbf{93.18$\pm$0.1} & 93.10$\pm$0.1 \\
15k       & \textbf{93.35$\pm$0.2} & 93.26$\pm$0.2 \\
16k       & 93.67$\pm$0.1 & \textbf{93.71$\pm$0.1} \\
17k       & \textbf{93.89$\pm$0.1} & 93.77$\pm$0.1 \\
18k       & \textbf{94.15$\pm$0.0} & 93.98$\pm$0.1 \\
19k       & \textbf{94.36$\pm$0.1} & 94.33$\pm$0.2 \\
20k       & \textbf{94.40$\pm$0.0} & \textbf{94.40$\pm$0.1}\\ \bottomrule
\end{tabular}
\caption{CIFAR-10 detailed results in training vs finetuning. This table corresponds to results of Figure 2b in the main paper.}
\label{supp:train_finetune}
\end{table*}
\begin{table*}[!thb]
\centering
\begin{tabular}{l|llllll}
\toprule
\#labeled & Random     & Divesity   & No diversity \\
\hline
1k        & \textbf{51.79$\pm$0.7} & 50.74$\pm$2.3 & 51.30$\pm$2.0   \\
2k        & 65.49$\pm$0.7 & \textbf{66.90$\pm$1.6} & 65.43$\pm$1.0   \\
3k        & 73.70$\pm$0.9 & \textbf{76.15$\pm$0.5} & 75.86$\pm$1.6   \\
4k        & 78.66$\pm$1.3 & \textbf{81.35$\pm$0.5} & 79.29$\pm$0.6   \\
5k        & 81.06$\pm$1.0 & \textbf{83.59$\pm$0.6} & 82.87$\pm$0.2   \\
6k        & 83.52$\pm$0.7 & \textbf{86.26$\pm$0.7} & 86.20$\pm$0.4   \\
7k        & 85.02$\pm$1.0 & \textbf{87.95$\pm$0.3} & 87.83$\pm$0.3   \\
8k        & 86.11$\pm$0.7 & \textbf{89.20$\pm$0.2} & 89.17$\pm$0.2   \\
9k        & 87.16$\pm$0.7 & \textbf{90.16$\pm$0.2} & 90.20$\pm$0.2   \\
10k       & 87.64$\pm$0.5 & \textbf{90.90$\pm$0.2} & 90.75$\pm$0.2   \\
11k       & 88.51$\pm$0.6 & \textbf{91.66$\pm$0.3} & 91.65$\pm$0.1   \\
12k       & 89.17$\pm$0.6 & \textbf{92.28$\pm$0.1} & 93.38$\pm$0.2   \\
13k       & 89.67$\pm$0.5 & \textbf{92.77$\pm$0.2} & 92.78$\pm$0.4   \\
14k       & 89.98$\pm$0.6 & \textbf{93.18$\pm$0.2} & 93.15$\pm$0.1   \\
15k       & 90.37$\pm$0.5 & \textbf{93.36$\pm$0.2} & 93.32$\pm$0.3   \\
16k       & 90.90$\pm$0.4 & \textbf{93.67$\pm$0.2} & 93.67$\pm$0.2   \\
17k       & 91.06$\pm$0.6 & \textbf{93.89$\pm$0.1} & 93.97$\pm$0.1   \\
18k       & 91.39$\pm$0.5 & \textbf{94.16$\pm$0.1} & 94.15$\pm$0.0   \\
19k       & 91.60$\pm$0.4 & \textbf{94.37$\pm$0.2} & 94.28$\pm$0.1   \\
20k       & 91.86$\pm$0.4 & \textbf{94.40$\pm$0.1} & 94.31$\pm$0.1\\ \bottomrule
\end{tabular}
\caption{CIFAR-10 detailed results in diversity. This table corresponds to results of Figure 3a in the main paper.}
\label{supp:diversity}
\end{table*}
\begin{table*}[!thb]
\centering
\begin{tabular}{l|llllll}
\toprule
\#labeled & Random     & Divesity   & No diversity \\
\hline
1k        & 49.98$\pm$1.2 & \textbf{49.98$\pm$1.8} & 49.98$\pm$1.4   \\
2k        & 65.10$\pm$1.0 & \textbf{65.70$\pm$1.4} & 58.72$\pm$1.0   \\
3k        & 70.99$\pm$1.0 & \textbf{74.00$\pm$0.8} & 66.45$\pm$0.7   \\
4k        & 78.2$\pm$1.0  & \textbf{80.59$\pm$0.8} & 72.93$\pm$0.9   \\
5k        & 79.19$\pm$0.8 & \textbf{82.79$\pm$0.7} & 75.63$\pm$0.7   \\
6k        & 82.92$\pm$0.7 & \textbf{84.96$\pm$0.7} & 79.98$\pm$0.6   \\
7k        & 84.88$\pm$0.8 & \textbf{87.31$\pm$0.5} & 81.32$\pm$0.6   \\
8k        & 85.84$\pm$0.6 & \textbf{89.07$\pm$0.4} & 83.61$\pm$0.5   \\
9k        & 87.08$\pm$0.5 & \textbf{89.66$\pm$0.4} & 85.04$\pm$0.4   \\
10k       & 87.28$\pm$0.4 & \textbf{90.36$\pm$0.3} & 86.04$\pm$0.4   \\
11k       & 88.48$\pm$0.5 & \textbf{90.87$\pm$0.4} & 86.79$\pm$0.4   \\
12k       & 88.58$\pm$0.4 & \textbf{91.78$\pm$0.2} & 87.83$\pm$0.4   \\
13k       & 89.26$\pm$0.4 & \textbf{92.08$\pm$0.3} & 88.01$\pm$0.4   \\
14k       & 89.85$\pm$0.4 & \textbf{92.7$\pm$0.3}  & 89.23$\pm$0.4   \\
15k       & 89.87$\pm$0.4 & \textbf{92.9$\pm$0.3}  & 89.27$\pm$0.4   \\
16k       & 90.17$\pm$0.4 & \textbf{93.36$\pm$0.3} & 90.35$\pm$0.3   \\
17k       & 90.69$\pm$0.4 & \textbf{93.56$\pm$0.3} & 90.47$\pm$0.3   \\
18k       & 91.04$\pm$0.3 & \textbf{93.82$\pm$0.2} & 90.98$\pm$0.3   \\
19k       & 90.99$\pm$0.3 & \textbf{93.94$\pm$0.2} & 91.63$\pm$0.3   \\
20k       & 91.28$\pm$0.3 & \textbf{93.95$\pm$0.2} & 91.53$\pm$0.3\\ \bottomrule
\end{tabular}
\caption{CIFAR-10x2 detailed results in diversity. This table corresponds to results of Figure 3b in the main paper.}
\label{supp:diversityx2}
\end{table*}
\begin{table*}[!thb]
\centering
\begin{tabular}{l|llllll}
\toprule
\#labeled & Random     & Divesity   & No diversity \\
\hline
1k        & 49.82$\pm$1.1 & \textbf{49.82$\pm$1.2} & 49.82$\pm$1.3   \\
2k        & 64.43$\pm$0.8 & \textbf{65.18$\pm$0.9} & 51.89$\pm$1.0   \\
3k        & 72.65$\pm$0.7 & \textbf{75.21$\pm$0.7} & 59.43$\pm$0.7   \\
4k        & 77.6$\pm$0.7  & \textbf{77.93$\pm$0.7} & 56.86$\pm$0.7     \\
5k        & 81.03$\pm$0.6 & \textbf{82.76$\pm$0.6} & 55.78$\pm$0.6   \\
6k        & 83.00$\pm$0.6 & \textbf{85.30$\pm$0.6} & 62.61$\pm$0.6   \\
7k        & 84.62$\pm$0.5 & \textbf{87.13$\pm$0.5} & 65.42$\pm$0.5   \\
8k        & 85.72$\pm$0.4 & \textbf{88.33$\pm$0.4} & 67.20$\pm$0.4   \\
9k        & 86.42$\pm$0.4 & \textbf{89.57$\pm$0.4} & 73.00$\pm$0.4   \\
10k       & 87.15$\pm$0.3 & \textbf{90.76$\pm$0.3} & 74.99$\pm$0.3   \\
11k       & 88.66$\pm$0.3 & \textbf{91.12$\pm$0.3} & 75.89$\pm$0.3   \\
12k       & 88.70$\pm$0.3 & \textbf{91.99$\pm$0.3} & 79.26$\pm$0.3   \\
13k       & 89.00$\pm$0.3 & \textbf{92.36$\pm$0.3} & 71.82$\pm$0.3   \\
14k       & 89.55$\pm$0.3 & \textbf{92.68$\pm$0.3} & 80.90$\pm$0.3   \\
15k       & 89.74$\pm$0.3 & \textbf{92.78$\pm$0.3} & 82.25$\pm$0.3   \\
16k       & 90.15$\pm$0.2 & \textbf{92.94$\pm$0.2} & 82.36$\pm$0.2   \\
17k       & 90.24$\pm$0.3 & \textbf{93.02$\pm$0.3} & 83.78$\pm$0.3   \\
18k       & 91.13$\pm$0.2 & \textbf{93.67$\pm$0.2} & 83.60$\pm$0.2   \\
19k       & 90.91$\pm$0.2 & \textbf{93.75$\pm$0.2} & 83.64$\pm$0.2   \\
20k       & 91.69$\pm$0.2 & \textbf{93.92$\pm$0.2} & 83.99$\pm$0.2\\ \bottomrule
\end{tabular}
\caption{CIFAR-10x5 detailed results in diversity. This table corresponds to results of Figure 3c in the main paper.}
\label{supp:diversityx5}
\end{table*}
\begin{table*}[!thb]
\centering
\begin{tabular}{l|llllll}
\toprule
\#labeled & Random     & Entropy    & LLAL       & Core-Set   & BALD       & Var Ratios \\
\hline
2k        & 60.95$\pm$0.2 & 61.23$\pm$0.6 & 60.95$\pm$0.6 & \textbf{62.36$\pm$0.5} & 61.23$\pm$0.4 & 60.89$\pm$0.5 \\
3k        & 64.18$\pm$0.2 & 64.57$\pm$0.4 & 64.91$\pm$0.5 & \textbf{65.90$\pm$0.5} & 64.64$\pm$0.3 & 64.24$\pm$0.3 \\
4k        & 66.39$\pm$0.2 & 66.94$\pm$0.2 & 66.90$\pm$0.3 & \textbf{67.63$\pm$0.2} & 66.92$\pm$0.2 & 67.12$\pm$0.2 \\
5k        & 67.46$\pm$0.3 & 68.70$\pm$0.2 & \textbf{69.05$\pm$0.2} & 68.88$\pm$0.2 & 68.12$\pm$0.2 & 68.04$\pm$0.2 \\
6k        & 68.58$\pm$0.4 & 69.82$\pm$0.3 & \textbf{70.35$\pm$0.2} & 69.74$\pm$0.2 & 68.95$\pm$0.2 & 68.87$\pm$0.2 \\
7k        & 69.17$\pm$0.2 & 70.48$\pm$0.2 & \textbf{70.49$\pm$0.2} & 70.16$\pm$0.2 & 69.43$\pm$0.1 & 69.26$\pm$0.2\\ \bottomrule
\end{tabular}
\caption{VOC07+12 detailed results. This table corresponds to results of Figure 4a in the main paper.}
\label{supp:voc07+12}
\end{table*}
\begin{table*}[!thb]
\centering
\begin{tabular}{l|llllll}
\toprule
\#labeled & Random     & Entropy    & Inconsistency \\
\hline
2k        & 63.12$\pm$0.2 & \textbf{63.18$\pm$0.2} & 63.15$\pm$0.2    \\
3k        & 67.08$\pm$0.1 & \textbf{67.44$\pm$0.1} & 67.42$\pm$0.1    \\
4k        & 69.44$\pm$0.1 & 70.22$\pm$0.1 & \textbf{70.42$\pm$0.1}    \\
5k        & 71.13$\pm$0.2 & \textbf{72.28$\pm$0.1} & 72.33$\pm$0.1    \\
6k        & 72.18$\pm$0.1 & \textbf{73.56$\pm$0.1} & 73.28$\pm$0.2    \\
7k        & 73.10$\pm$0.1 & 74.81$\pm$0.1 & \textbf{74.85$\pm$0.1}\\ \bottomrule
\end{tabular}
\caption{VOC07+12 detailed results in semi-supervised learning. This table corresponds to results of Figure 4b in the main paper.}
\label{supp:cons_voc07+12}
\end{table*}
\begin{table*}[!thb]
\centering
\begin{tabular}{l|llllll}
\toprule
\#labeled & Random     & Entropy    & Inconsistency \\
\hline
250       & 87.58$\pm$0.2 & 88.85$\pm$0.2 & \textbf{89.97$\pm$0.3}    \\
500       & 90.37$\pm$0.1 & 90.89$\pm$0.2 & \textbf{91.57$\pm$0.2}    \\
1k        & 91.76$\pm$0.1 & 92.47$\pm$0.2 & \textbf{92.87$\pm$0.1}    \\
2k        & 92.78$\pm$0.1 & 93.27$\pm$0.2 & \textbf{93.74$\pm$0.2}    \\
4k        & 93.38$\pm$0.1 & 93.97$\pm$0.1 & \textbf{94.23$\pm$0.1}\\ \bottomrule
\end{tabular}
\caption{CIFAR-10 detailed results in semi-supervised learning. This table corresponds to results of Figure 4c in the main paper.}
\label{supp:cons}
\end{table*}

\end{document}